\documentclass[11pt]{article}

\usepackage[final]{acl}

\PassOptionsToPackage{hyphens}{url}\usepackage{hyperref}

\usepackage{times}
\usepackage{latexsym}

\usepackage[T1]{fontenc}

\usepackage[utf8]{inputenc}

\usepackage{microtype}

\usepackage{inconsolata}

\usepackage{graphicx}

\usepackage{xurl}
\usepackage{hyperref}
\usepackage{booktabs}
\usepackage{amsfonts}
\usepackage{nicefrac}
\usepackage{microtype}
\usepackage{xcolor}
\usepackage[para]{footmisc}

\usepackage{graphicx}
\usepackage{amsmath}
\usepackage{amssymb}
\usepackage{amsthm}
\usepackage{subcaption}
\usepackage{todonotes}
\usepackage{enumitem}
\usepackage{listings}
\usepackage{siunitx}
\usepackage{multirow}
\usepackage{tabularx}
\usepackage{makecell}
\usepackage{tabulary}

\usepackage{threeparttable}
\usepackage{courier}
\usepackage{longtable}
\usepackage{geometry}
\usepackage{pdflscape}

\usepackage{pifont}
\newcommand{\cmark}{\ding{51}}%
\newcommand{\xmark}{\ding{55}}%
\usepackage{siunitx}
\usepackage{dcolumn}
\newcolumntype{d}[1]{D{.}{.}{#1}}

\title{ScreenQA: Large-Scale Question-Answer Pairs \\Over Mobile App Screenshots}

\author{
  Yu-Chung Hsiao\thanks{Co-first authors with equal contributions.}\thanks{This work was done while the author was at Google.}\qquad
  Fedir~Zubach\footnotemark[1]\qquad
  Gilles Baechler\qquad
  Srinivas Sunkara
  \\
  \textbf{Victor C\u{a}rbune\qquad
  Jason Lin\qquad
  Maria Wang\qquad
  Yun Zhu\qquad
  Jindong~Chen\thanks{Corresponding author.}} \vspace*{5px}\\
  Google DeepMind \qquad Cisco Systems\footnotemark[2] \vspace*{5px}\\
  \normalsize{\texttt{yohsiao@cisco.com, \{fedir, baechler, srinivasksun,}}\\
  \normalsize{\texttt{vcarbune, jasonjlin, mariawang, yunzhu, jdchen\}@google.com}} \\
}

\begin{document}
\maketitle
\setlength{\abovedisplayskip}{3pt}
\setlength{\belowdisplayskip}{3pt}

\begin{abstract}
We introduce ScreenQA, a novel benchmarking dataset designed to advance screen content understanding through question answering. The existing screen datasets are focused either on low-level structural and component understanding, or on a much higher-level composite task such as navigation and task completion for autonomous agents. ScreenQA attempts to bridge this gap. By annotating~86k question-answer pairs over the RICO dataset, we aim to benchmark the screen reading comprehension capacity, thereby laying the foundation for vision-based automation over screenshots.
Our annotations encompass full answers, short answer phrases, and corresponding UI contents with bounding boxes, enabling four subtasks to address various application scenarios. We evaluate the dataset's efficacy using both open-weight and proprietary models in zero-shot, fine-tuned, and transfer learning settings. 
We further demonstrate positive transfer to web applications, highlighting its potential beyond mobile applications.
\end{abstract}

\section{Introduction}
\label{sec:introduction}

\begin{figure*}[t]
    \centering
        \begin{subfigure}[t]{0.3\textwidth}
        \centering
        \includegraphics[width=\textwidth]{figures/rico_app_honestly_0_457.png}
        \caption{Question: ``How many likes and comments are there for the post \textit{Why Michael Flynn ...}?''}
        \label{fig:honestly}
    \end{subfigure}  \hfill
    \begin{subfigure}[t]{0.3\textwidth}
        \centering
        \includegraphics[width=\textwidth]{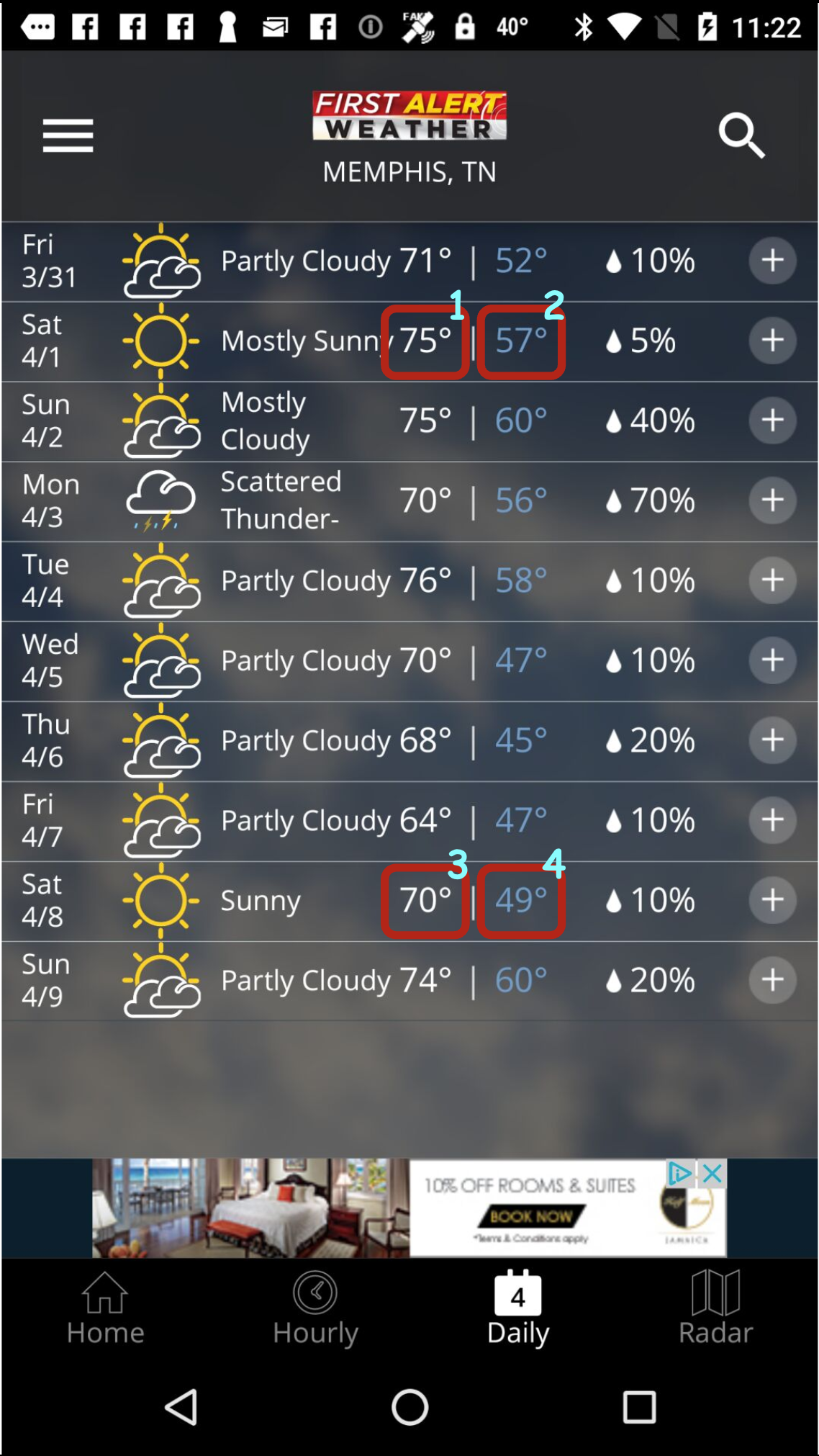}
        \caption{Question with Ambiguity: ``What's the temperature on Saturday?''}
        \label{fig:eg-weather-forecast}
    \end{subfigure}  \hfill
    \begin{subfigure}[t]{0.3\textwidth}
        \centering
        \includegraphics[width=\textwidth]{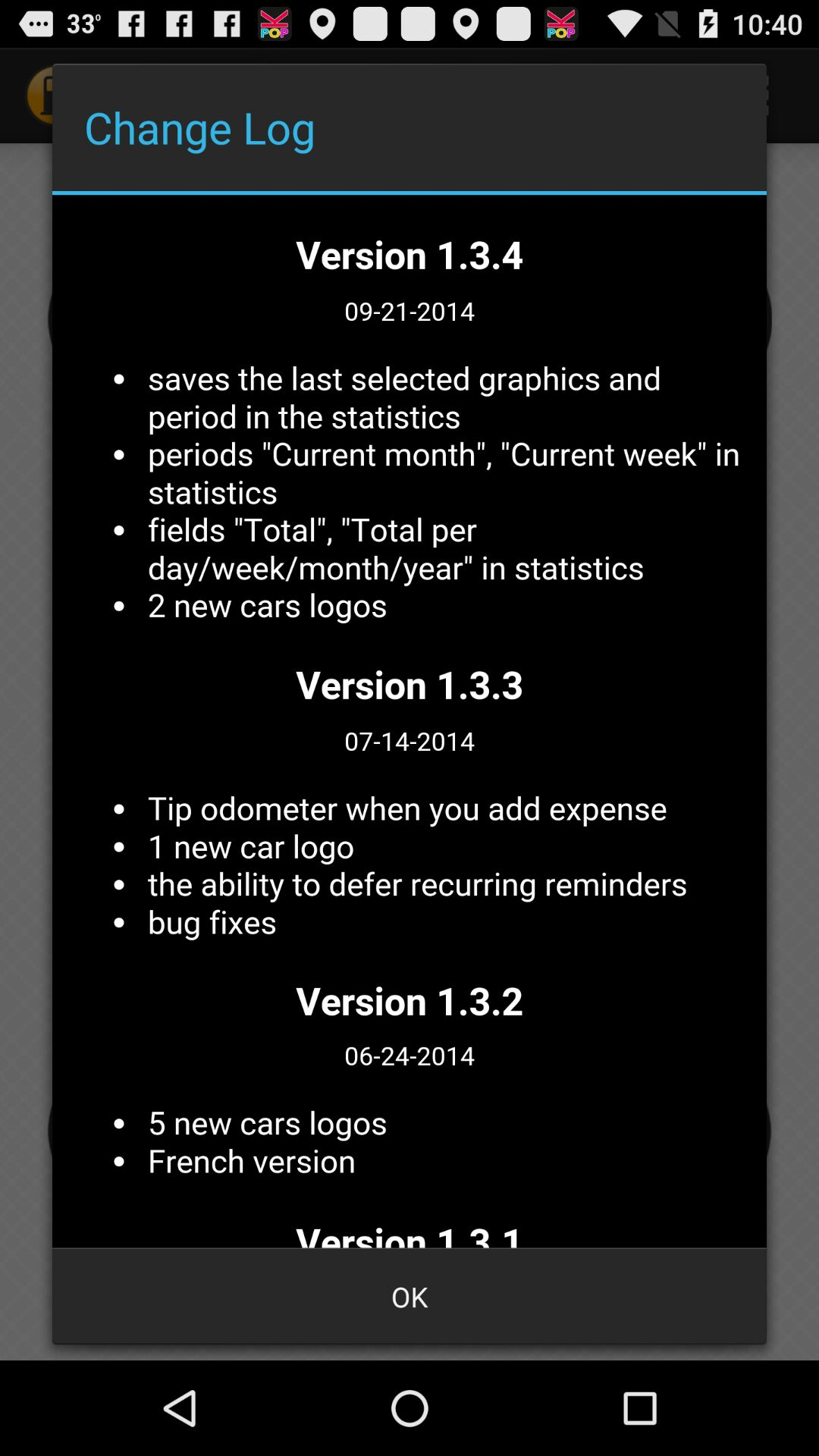}
        \captionsetup{belowskip=10pt}
        \caption{No answer to the question: ``What is the date of version 1.3.1?''}
        \label{fig:no_answer_date}
    \end{subfigure}\hfill

    \caption{
        ScreenQA examples. 
        (a) Four corresponding tasks (Section~\ref{sec:tasks_metrics}):
        1) Short answer (SQA-S): "1,1". 2) Long answer (SQA-L): "There is 1 like and 1 comment". 3) UI Content (SQA-UIC): [``1'', ``1'']. 4) UI Content w/ Bounding Boxes (SQA-UIC-BB): also with bounding box coordinates (in green). (b) Both high and low temperatures in two Saturdays are relevant answers. (c) Lacking the content to answer the question.
    }
    \label{fig:screenqa-task-example}
\end{figure*}

Recent advancements in machine learning, especially Visual Large Language Models (VLMs) or Multimodal LLMs (MLLMs), have catalyzed numerous applications centered on mobile screens. 
These applications range from personal assistants enabling hands-free or eyes-free interaction to code generation from UI design mock-ups, adaptive device interfaces, automated ad creation, and mobile app testing.
All of these require reliable screen content understanding as the foundation to ensure quality applications.

Mobile app screenshots have been analyzed using machine learning from multiple perspectives.
These include pixel-level understanding, such as layout structural analyses, UI issue detection and correction~\cite{liLearningDenoiseRaw2022}, UI element semantics like icon recognition, button action prediction~\cite{sunkaraBetterSemanticUnderstanding2022}, to even higher-level functional analyses such as accessibility support~\cite{liWidgetCaptioningGenerating2020}, screen description~\cite{wangScreen2WordsAutomaticMobile2021}, and screen type classification~\cite{dekaRicoMobileApp2017}.
However, the specific aspect of understanding screen contents from pixels remains comparatively understudied.

Screen contents encompass diverse information, from restaurant ratings to system settings and chat messages. 
This capacity is crucial because:
1) screens primarily function to present information, and
2) autonomous agents and task completion systems require precise screen content understanding
to achieve reasonable success rates for multi-step processes.


In this work, we propose using pixels as the sole representation of UI screens, to eliminate dependency on unreliable structural representations such as view hierarchies (VHs)~\cite{zang2021multimodal} and maximize applicability to various visual understanding tasks. 
To this end, we annotated the RICO dataset~\cite{dekaRicoMobileApp2017} with~85,984 question-answer pairs (Section~\ref{sec:data_annotation}) and defined four tasks with evaluation metrics (Section~\ref{sec:tasks_metrics}), creating {\it ScreenQA}\footnote{ScreenQA dataset is released at \href{https://github.com/google-research-datasets/screen_qa}{\texttt{https://github.com}}\\ \href{https://github.com/google-research-datasets/screen_qa}{\texttt{/google-research-datasets/screen\_qa}}}.

RICO remains the preferred screenshot dataset due to its
1) extensive collection of 66k screenshots, 2) diverse app distribution spanning 9.3k apps across~27 categories, 
3) numerous benchmarking tasks built upon RICO that enable cross-referencing and comparison\footnote{We deliberately chose not to combine RICO with other datasets for this reason.}~\cite{wangScreen2WordsAutomaticMobile2021, liWidgetCaptioningGenerating2020, liLearningDenoiseRaw2022, ahmed_masc_2023, luAIAssistanceUX2024a}, and 4) widespread use as a training dataset in recent document/screen understanding and VLM research~\cite{chengSeeClickHarnessingGUI2024a, liuTextMonkeyOCRFreeLarge2024a, xie_learning_2023}.
Despite the availability of newer datasets, RICO's comprehensive nature and established presence in the field make it an optimal choice for this work. See Table~\ref{tab:rico_comparison} for comparison.

This work offers the following contributions:

\begin{enumerate}[nolistsep, leftmargin=*]
    \item We create and release ScreenQA, the first large-scale question answering dataset and benchmark for mobile screens.
    \item ScreenQA provides rich annotations: short answers, full-sentence long answers, and 
    supporting UI element contents with their bounding boxes (Section~\ref{sec:data_annotation}).
    \item We propose four tasks with evaluation metrics to satisfy various application needs in extraction, text generation, and grounding (Section~\ref{sec:tasks_metrics}).
    \item We establish zero-shot, fine-tuning and cross-domain transfer learning baselines for both closed and open-weight models quantifying the effectiveness of ScreenQA (Section~\ref{sec:baselines}).
\end{enumerate}


\begin{table}[ht]
    {
        \centering
        \scriptsize
        \setlength{\tabcolsep}{4pt}
        \begin{threeparttable}
        \begin{tabular}{lrlrr}
            \toprule
            Dataset & Size & Unit & Apps & Categories \\
            \midrule
            RICO~\cite{dekaRicoMobileApp2017} & 66k & screenshot & 9.3k & 27 \\
            LabelDroid~\cite{chen2020unblind} & \underline{13k} & screenshot & 7.6k & 25 \\
            MoTIF~\cite{burns2022motifvln} & \underline{4.7k} & trace & \underline{0.1k} & \underline{n/a} \\
            AitW~\cite{rawlesAndroidWildLargescale2024} & 715k & trace & \underline{0.4k} & \underline{n/a} \\
            \bottomrule
        \end{tabular}
        \end{threeparttable}
    }\\
    \caption{The RICO dataset remains the most suitable mobile app screenshot collection for this work due to its diversity, large scale, and established presence in the field. 
    Although AitW is larger in scale, it lacks app and screenshot diversity due to its sampling method.
    } 
    \label{tab:rico_comparison}    
\end{table}

\section{Related Work}
\label{sec:related_work}

Closest to ScreenQA is work that tackles multimodal understanding and question answering, as discussed below.



\subsection{Multimodality}
We discuss the research on text-heavy images\footnote{Natural image VQA and video VQA are omitted due to the space constraints.} and provide an overview in Table~\ref{tab:dataset_comparison}.

\begin{table*}[ht]
    \centering{
        \small
        \setlength{\tabcolsep}{4pt}
        \begin{threeparttable}
        \begin{tabular}{lS[table-number-alignment = center]S[table-number-alignment = center]cl}
            \toprule
            Dataset & \multicolumn{1}{c}{\# Images (k)} & \multicolumn{1}{c}{\# Examples (k)} & UI? & Task Type \\
            \midrule
            ScreenQA (this work)\tnote{*} & 35 & 86 & \cmark & Question Answering (QA) \\
            \cmidrule{1-5}
            TextVQA~\cite{singh2019textvqa} & 28 & 45 & \xmark & QA \\
            DocVQA~\cite{mathew2021docvqa} & 12 & 50 & \xmark & QA \\
            InfographicVQA~\cite{mathew2022infographicvqa}	& 5.5 & 30 & \xmark & QA \\
            ChartQA~\cite{masry2022chartqa}	& 22 & 33 & \xmark & QA \\
            WebSRC~\cite{chenWebSRCDatasetWebBased2021} & 6.5 & 440 & \cmark & QA on Webpage Segments \\            
            ComplexQA~\cite{baechler2024screenai} & 10 & 12 & \cmark & QA on Counting, Arithmetic \\
            VisualWebBench - WebQA~\cite{liuVisualWebBenchHowFar2024} & 0.3 & 0.3 & \cmark & QA on Webpages -- Eval Only \\
            LabelDroid~\cite{chen2020unblind} & 13 & 19 & \cmark & Text Generation \\            
            Screen2Words~\cite{wangScreen2WordsAutomaticMobile2021} & 22 & 112 & \cmark & Summarization \\
            ScreenAnnotation~\cite{baechler2024screenai} & 22 & 22 & \cmark & Object Detection \\
            MoTIF~\cite{burns2022motifvln} & 62 & 4.7 & \cmark & Navigation \\
            AitW~\cite{rawlesAndroidWildLargescale2024}\tnote{$\dagger$} & 5690 & 715 & \cmark & Navigation \\
            \bottomrule
        \end{tabular}
        \begin{tablenotes}
            \item[*] A subset of the 66k images from RICO~\cite{dekaRicoMobileApp2017} as described in Section~\ref{sec:data_annotation}
            \item[$\dagger$] Limited app and screenshot diversity as described in Table~\ref{tab:rico_comparison}
        \end{tablenotes}        
        
        \end{threeparttable}
    }\\
    \caption{
    Comparison of ScreenQA with Related Datasets. ScreenQA is the largest QA dataset for mobile screenshots, using entire screenshots rather than cropped answer regions (as in WebSRC) to better represent real-world applications, and including unanswerable questions and bounding boxes. ComplexQA complements ScreenQA with its focus on counting, arithmetic, and comparisons. Not an exhaustive comparison due to space constraints.
    } 
    \label{tab:dataset_comparison}    
\end{table*}

\paragraph{Screen UI-based understanding}
Unlike natural images, UIs are designed to be informative and actionable. 
Examples of work dealing with informativeness include 1) UI element identification, such as icon detection~\cite{dekaRicoMobileApp2017}, widget captioning~\cite{liWidgetCaptioningGenerating2020, chen2020unblind}, and 2) referring expression to UI elements in  classification~\cite{wu2023webui}, representation learning~\cite{bai2021uibert} and generation~\cite{hong2023cogagent}. 
Actionability is related to task completion and autonomous agents.
MoTIF \cite{burns2022motifvln}, VisualWebArena~\cite{koh2024visualwebarena},
and
Android-in-the-Wild~\cite{rawlesAndroidWildLargescale2024} provide interactive app environments for evaluating visually grounded screen agents. 
This work is focused on the information retrieval aspect.

\paragraph {Visual Document Understanding}
Visual Document Understanding concerns understanding scanned or photographed documents.
DocVQA~\cite{mathew2021docvqa} uses an extractive QA format for span/segment extraction. 
TextVQA~\cite{singh2019textvqa} and several other domain-specific datasets \cite{mishra2019ocrvqa, gurariVizWizGrandChallenge2018a, huangICDAR2019CompetitionScanned2019a, abdallahCORUComprehensivePostOCR2024, jaumeFUNSDDatasetForm2019a, harleyEvaluationDeepConvolutional2015a, lewisBuildingTestCollection2006a} 
relate the~2D arrangement of texts to semantic meanings.
Beyond text alone, infographics understanding~\cite{mathew2022infographicvqa, masry2022chartqa, kahou2017figureqa, kafleDVQAUnderstandingData2018a, chaudhryLEAFQALocateEncode2020a} focus on charting with text around.



\subsection{Question Answering}
We focus on closed-domain question answering grounded in specific contexts.
Answer formats include span~\cite{rajpurkarSQuAD1000002016a}, entity~\cite{talmorWebKnowledgeBaseAnswering2018a}, multiple choice~\cite{mihaylovCanSuitArmor2018a}, and generation~\cite{xiongTWEETQASocialMedia2019a}. 
Capacities range from reading comprehension~\cite{yangWikiQAChallengeDataset2015a}, multi-hop reasoning~\cite{yangHotpotQADatasetDiverse2018a},~\cite{chenFinQADatasetNumerical2021a}, logical reasoning~\cite{yuReClorReadingComprehension2020a}, and commonsense reasoning~\cite{talmorCommonsenseQAQuestionAnswering2019a}.

\section{Problem Setting: Tasks and Metrics}
\label{sec:tasks_metrics}


As each example in the validation and test sets may contain alternative ground truths from multiple annotators, we compute the evaluation metric by: 1) calculate the maximum metric value across all annotator-provided ground truths for each example, and 2) average these maximum values across the entire dataset:
\[
\textrm{avg}(metric) = \frac{1}{N}\sum_{i=1}^N \max_j [ metric(A_i, A^g_{i,j})], 
\]
where $N$ is the number of questions, $A_i$ is the predicted answer for $i$-th question, and $A^g_{i,j}$ is the $j$-th ground truth for $i$-th question.

ScreenQA (or SQA in short) presents four tasks encompassing diverse application scenarios.
Examples are illustrated in Figure~\ref{fig:screenqa-task-example}.

\paragraph{ScreenQA Short answer (SQA-S)}
Given a screenshot and a question, output a short (concise) answer to this question using the information presented on the screen.
If the screenshot doesn't contain the answer, output ``\(<\)no answer\(>\)''.


Similar to SQuAD~\cite{rajpurkarSQuAD1000002016a}, we propose to use \textit{Exact Match} (EM) and
\textit{F1-Score} for accommodating acceptable permutations and rephrasing of the same content.
We also apply SQuAD pre-processing to normalize answers before computing the metrics.
This task is the core capability enabled by the dataset.

\paragraph{ScreenQA Long answer (SQA-L)}
This task is identical to SQA-S, with the distinction that it requires a long, full-sentence answer instead of short answer phrases.
This task facilitates the generation of coherent responses suitable for direct human interaction, particularly in the context of virtual assistant applications. 

As the task resembles summarization of pertinent information, we propose to use \textit{ROUGE-\{1,2,L\}}~\cite{lin-2004-rouge} as the evaluation metric.

\paragraph{ScreenQA UI Content (SQA-UIC)}
Given a screenshot and a question, output a list of UI elements that contain the answer, where each element is represented by its text representation.
If the screenshot doesn't contain the answer, output an empty list.

Section~\ref{sec:data_annotation} will further details the UI elements corresponding to questions, their listed order, and their contents represented in text.
Except for icons with predefined textual descriptions, most content resembles OCR output: text within UI elements~\cite{Qin_2019_ICCV}. However, unlike OCR, the output should be evaluated as a list rather than a continuous sequence of symbols or words.


We use \textit{Exact Match} and \textit{F1-score} as metrics.
Commonly, text in screenshots is directly extractable. Therefore, we perform UI-element-wise matching without additional pre-processing.

\paragraph{ScreenQA UI Content with Bounding Boxes (SQA-UIC-BB)}
Given a screenshot and a question, output a list of UI elements that contain the answer, where each element is represented by its bounding box \textit{and} text representation.

This task extends the previous SQA-UIC, additionally facilitating answer highlighting and action automation over the screen. 
The detection of bounding boxes, especially in screen contents, is rarely available in existing datasets and challenges model capabilities.



We recommend evaluating the bounding box detection quality using \textit{F1-Score}, where two bounding boxes match if their Intersection over Union (IoU) \cite{rezatofighi2019generalized} score is higher than~$0.1$.
\textit{Exact match} and \textit{F1-score} are evaluated in a similar way, but restricting the list of matches to only those where text representation also matches.
This threshold is justified due to the annotation methods of ground truth bounding boxes: selection from view hierarchies (accommodating UI elements with substantial no-content area) or manually drawing (tightly fitting text) (Section~\ref{sec:answer_annotation}).
The simultaneous use of these approaches by different annotators typically results in low IoU values.


\section{Data Annotation}
\label{sec:data_annotation}

\begin{table}[t]
\centering
    {
        \footnotesize
        \setlength{\tabcolsep}{4pt}
        \begin{threeparttable}
        \begin{tabular}{llrr}
            \toprule
            Stage & Step & \# SS & \# Q\\
            \midrule
            & RICO Original & 66k & \multicolumn{1}{r}{--} \\
            \cmidrule{1-4}
            Prefiltering  & Non-English apps\tnote{$\dagger$} & 11k &  \multicolumn{1}{r}{--} \\
            Section~\ref{sec:pre-filtering} & Out-of-sync VHs\tnote{$\dagger$} & 13k &  \multicolumn{1}{r}{--} \\
            \cmidrule{2-4}
            & For Question Anno. & 51k &  \multicolumn{1}{r}{--} \\
            \cmidrule{1-4}
            Question Anno. & 1st pass of Q anno. & \multirow{2}{*}{35k} & 46k \\
             Section~\ref{sec:question_annotation} & 2nd pass of Q anno. & & 36k \\
            & Lack of content & 15k &  \multicolumn{1}{r}{--} \\
            \cmidrule{2-4}
            & For Answer Anno. & 35k & 82k \\
            \cmidrule{1-4}
            \multirow{2}{*}{\shortstack[l]{Not Ans. Anno.\\ Section~\ref{sec:not_answerable_question_annotation}}} & 3rd pass of Q anno. & 5k & 5k \\
            \cmidrule{2-4}
             & For Data Splitting & 35k & 86k \\
            \cmidrule{1-4}
            Data Splitting & Train & 28,378 & 68,951 \\
            Section~\ref{sec:dataset_statistics} & Validation & 3,485 & 8,614 \\
            & Test & 3,489 & 8,419 \\
            \cmidrule{2-4}
            & Total & 35,352 & 85,984 \\
            \bottomrule
        \end{tabular}
        \begin{tablenotes}
        \item[$\dagger$] Not mutually exclusive.
        \end{tablenotes}        
        \end{threeparttable}
    }
    \caption{Counts of distinct screenshot (SS) and questions (Q) for annotation stages and data splitting.} 
    \label{tab:prefiltering_numbers}    
\end{table}

We performed a five-stage process to annotate ScreenQA: 1) Prefiltering~(Section~\ref{sec:pre-filtering}), 2) Question annotation~(Section~\ref{sec:question_annotation}), 3) Answer annotation~(Section~\ref{sec:answer_annotation}), 4) Not-answerable question annotation~(Section~\ref{sec:not_answerable_question_annotation}), and 5) Short answer generation~(Section~\ref{sec:short_answers_generation}).
See data statistics in Table~\ref{tab:prefiltering_numbers}, data annotation details and UI tools in  Appendix~\ref{appendix:annotation_ui}, and data examples in Appendix~\ref{appendix:data_examples}.

\subsection{Prefiltering}
\label{sec:pre-filtering}

Prefiltering involves one round of human rating to remove screenshots with the following conditions: 
\begin{enumerate}[nolistsep, leftmargin=*]
\item Screenshots from non-English apps. 
\item Screenshots with unsynchronized view hierarchies (VHs).
\end{enumerate}
Removing these data helps avoid language-related obstacles and potential view hierarchy issues~\cite{zang2021multimodal}, which could otherwise introduce noise into subsequent annotation stages.
We employed~27 annotators to perform this stage.
Occlusion and ghosting during screen transitions are deemed acceptable if the UI elements in the main content area remain clean and accurate, resulting in different numbers from~\cite{liMappingNaturalLanguage2020a}.
Examples of such VH symptoms are provided in Appendix~\ref{appendix:prefiltering_rules}.

\subsection{Question Annotation}
\label{sec:question_annotation}

We employed the same~27 annotators to formulate questions based on provided screenshots, simulating real-world app usage queries. 
Questions were restricted to directly observable information on the screen, deliberately excluding those questions requiring reasoning, computation, or advertisement-related content.
This is because recent multimodal reasoning benchmarks~\cite{yueMMMUMassiveMultidiscipline2024, hanInfiMMEvalComplexOpenEnded2023} unveil that the multimodal LLM reasoning capacities are heavily influenced by their underlying LLM performance (Section 6 of~\cite{wangExploringReasoningAbilities2024}), which can be more efficiently achieved through inference time scaling~\cite{brownLargeLanguageMonkeys2024, snellScalingLLMTestTime2024, wuEmpiricalAnalysisComputeOptimal2024, kumarTrainingLanguageModels2024}, followed by pretraining data and model parameter scaling~\cite{zhangWhenScalingMeets2024, owenHowPredictableLanguage2024}.
By focusing our scope on screen content understanding, we aim to investigate multimodality independent of testing other types of intelligence.

Question annotation was a two-pass process.
First, an annotator composed up to five questions per screenshot.
Then, a second annotator, with access to the initial questions, added up to three more, resulting in a maximum of eight questions per screenshot.
This two-pass approach promoted a diverse and comprehensive set of questions, while reducing redundancies.
To encourage diverse questioning styles and a varied set of questions, each annotator was required to complete their own initial set of questions before they could see questions created by others.
Screenshots lacking interesting content (e.g., login pages, ads) may be
skipped.


\subsection{Answer Annotation}
\label{sec:answer_annotation}

Given a screenshot and a question, the annotators are tasked to
\begin{enumerate}[nolistsep, leftmargin=*]
    \item Correct grammatical errors in the question.
    \item Answer the question by:
        a) Annotating bounding boxes that constitute the answer.
        b) Ranking bounding boxes by relevance or reading order.
    \item Provide a full-sentence long answer.
\end{enumerate}
The annotator who composed a specific question is excluded from answering it to prevent potential bias.
Our UI tool offers two bounding box annotation methods: direct selection of UI elements from view hierarchy(VH) leaf nodes, or manual drawing when VH structure is inadequate for selection.

We further consider two exceptions: 1) Mark ``invalid question'' if incomprehensible. 2) Mark ``not answerable from the screenshot'' if the screenshot does not contain the answer. 
The ``invalid question'' answer annotations are first removed for each question.
Then questions without any remaining answer annotations are excluded from the ScreenQA dataset, as they are deemed invalid.
To enhance evaluation quality, three answer were annotated for validation and test sets, while only one for training set (details in Section~\ref{sec:dataset_statistics}).

\subsection{Not-Answerable Question Annotations}
\label{sec:not_answerable_question_annotation}

\begin{figure}[t]
    \centering
    \includegraphics[width=1.02\columnwidth]{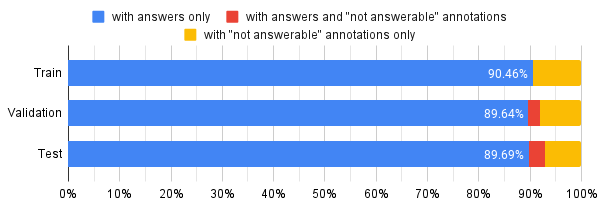}
        \caption{
            Distribution of question answerability. Each split contains approximately~10\% questions that have at least one ``not answerable'' answer annotation, to test models' ability to refrain from answering unanswerable questions.
        }
        \label{fig:q_status}
\end{figure}

A third pass of question annotation was implemented to increase ``not-answerable questions'' to approximately 10\%~(Figure~\ref{fig:q_status}).
This tests models' ability to refrain from answering unanswerable questions, similar to~\cite{rajpurkarKnowWhatYou2018a}.
Approximately 5k screenshots without existing ``not answerable'' questions were randomly sampled. 
We instructed annotators to compose one question per screenshot that was related to the screen content but unanswerable.
See an example in Figure~\ref{fig:no_answer_date}.


\subsection{Short Answer Generation}
\label{sec:short_answers_generation}

As UI design of apps may fragment a short, semantic answer into multiple bounding boxes (e.g., ``Oct. 15, 2024'' fragmented into ``Oct.'', ``15'', and ``2024'' in a date selector), we perform a post-processing step to make the annotations into a coherent, human readable short answer suitable for common question answering purposes.
This also partly accommodate answer variants that are semantically identical, such as numerical representations (digits/words), date/time formats, and optional descriptors/units (e.g., ``3 reviews'' vs. ``3'').
We employed PaLM~2~\citep{anil2023palm} to generate answer variants via few-shot prompting with inputs including question, UI element descriptions, and full-sentence answer, with a subset verified by authors.
See Appendix~\ref{appendix:screenqa_short_prompts} for the prompt details.




\section{Dataset Analysis}
\label{sec:dataset_analysis}

\begin{figure*}[t]
    \centering
    \begin{subfigure}[t]{0.48\textwidth}
        \centering
        \includegraphics[width=\textwidth, height=1.5in]{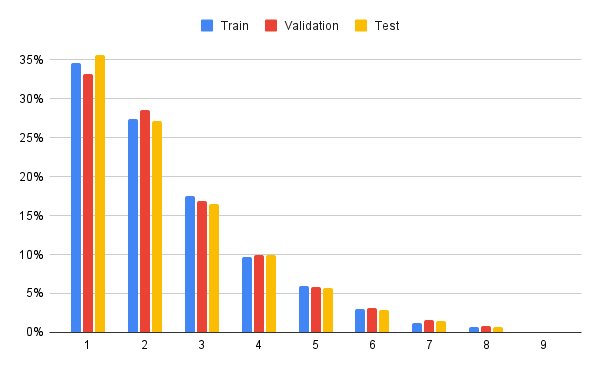}
        \caption{
            Composed questions per screenshot.
        }
        \label{fig:q_num}
    \end{subfigure}\hfill
    \begin{subfigure}[t]{0.48\textwidth}
        \centering
        \includegraphics[width=\textwidth, height=1.5in]{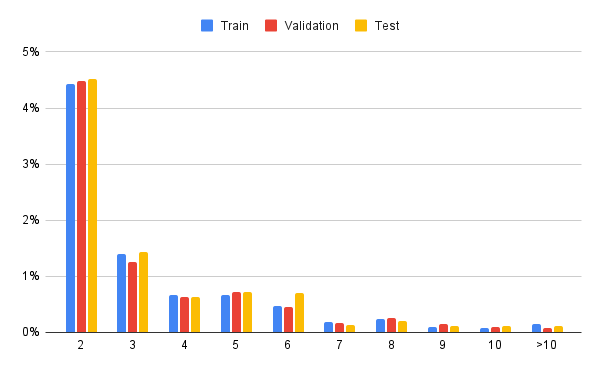}
        \caption{
            Bounding boxes used to answer a question.
        }
        \label{fig:a_num}
    \end{subfigure}\hfill
    \begin{subfigure}[t]{0.48\textwidth}
        \centering
        \includegraphics[width=\textwidth]{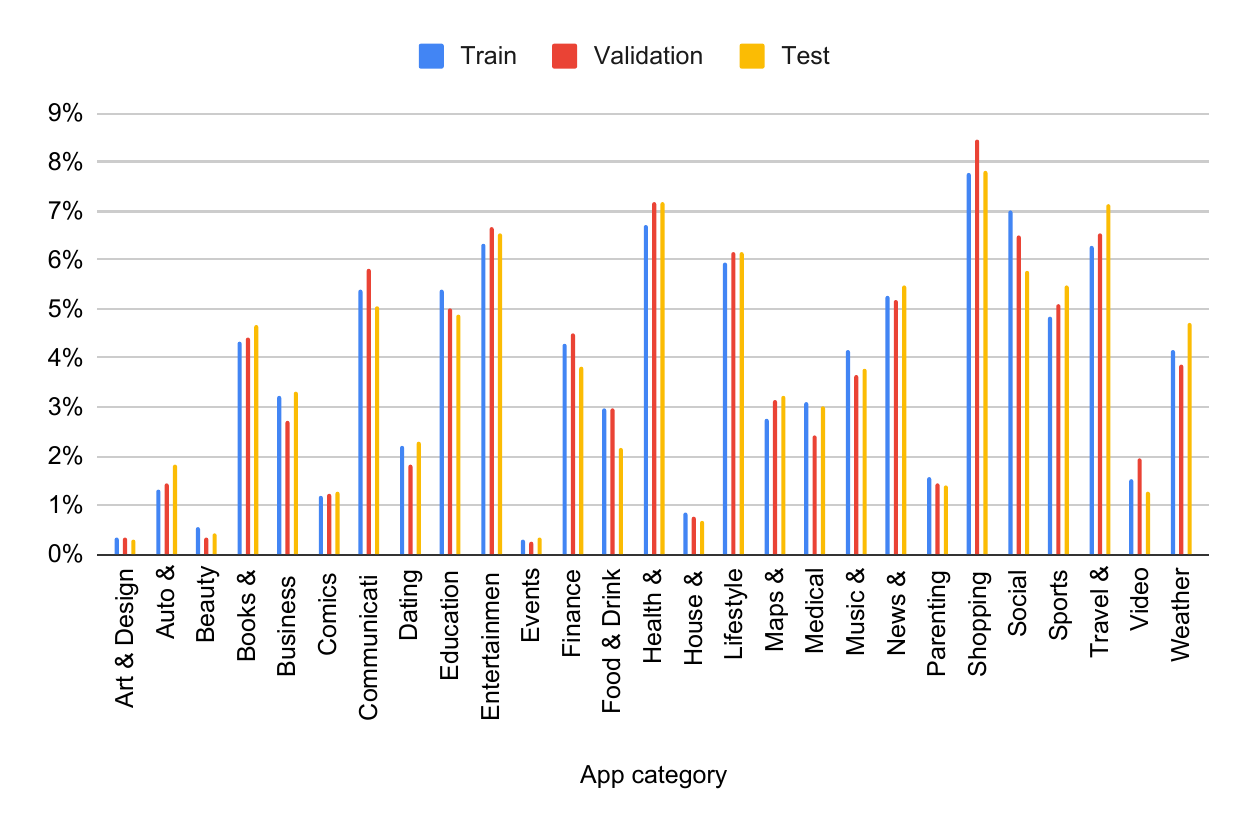}
        \caption{
            Questions per app category.
        }
        \label{fig:q_per_app_category}
    \end{subfigure}\hfill
    \begin{subfigure}[t]{0.48\textwidth}
        \centering
        \includegraphics[width=\textwidth]{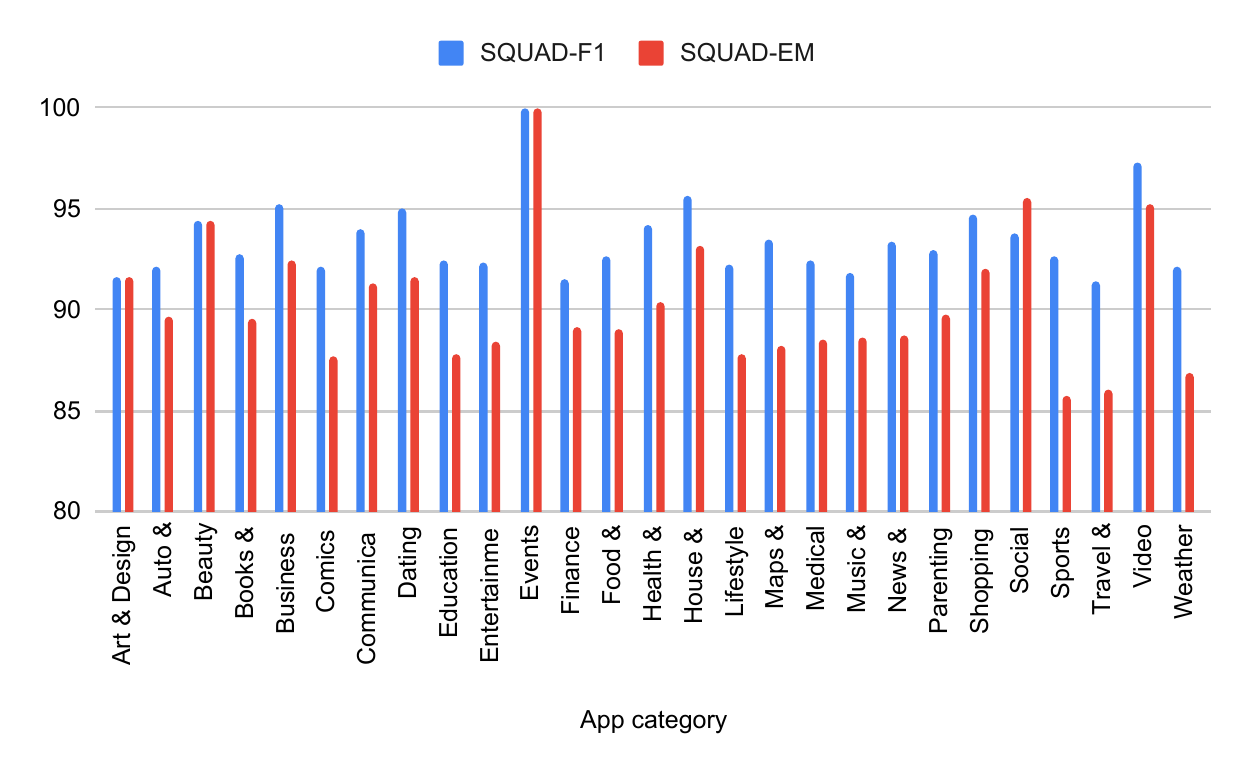}
        \caption{
            Fine-tuned PaliGemma 896 performance on SQA-S.
        }
        \label{fig:sqa_s_metric_per_category}
    \end{subfigure}\hfill
    \caption{
        Histograms for various data and model performances.
        (b) Approximately~91\% of questions are either not answerable or can be answered by a single bounding box, hence, omitted to emphasize the long tail distribution. 
        (d) The model exhibits consistent SQuAD-F1 performance across app categories, except for Events and Videos, which present higher scores due to lower support for those categories.
    }
    \label{fig:qa_histograms}
\end{figure*}

\subsection{Dataset Splitting}
\label{sec:dataset_statistics}

The ScreenQA dataset comprises 85,984 questions from 35,352 unique screenshots, split at the screenshot level into training, validation, and test sets in an approximate 80-10-10 ratio (see Table~\ref{tab:prefiltering_numbers}).


\subsection{Question Analysis}
\begin{table*}[t]
    \centering
    \scriptsize
    \begin{tabular}{lrl|l}
        \toprule
        Category & \% & \multicolumn{2}{l}{Examples} \\
        \cmidrule(lr){1-2} \cmidrule(lr){3-4}
        UI selection \& config & 18.1 & Which option is selected? & What is the selected ringtone? \\
        Quantity number & 11.7 & How many unread messages? & How many pictures are there in Western Europe? \\
        App name & 10.4 & What is the name of the application? & What is the app name? \\
        Date time & 9.4 & When was ``Heal the Living'' released? & When is happy hour? \\
        Price & 3.4 & How much is the gift bonus in 3rd place? & What is the price? \\
        Name of item & 3.3 & What is the name of the drug? & What is the name of chef? \\
        User name & 2.8 & What is the name of the user? & What is the username on telegram? \\
        Duration & 2.5 & What is the duration of video? & How long is the song? \\
        Enum. of avail. options & 2.5 & Which social media options are given there? & What are the options available for logging in? \\
        Address and direction & 2.4 & What is the current location? & What is the service zip code? \\
        Email address & 2.4 & What is an email address? & What is customer service email? \\
        Person's name & 2.1 & Who sang the song? & What is the last name? \\
        \cmidrule(lr){1-2} \cmidrule(lr){3-4}
        Others & 12.8 & \multirow{2}{*}{\shortstack[l]{What's the average speed?\\ What is the spending limit?}} & \multirow{2}{*}{\shortstack[l]{What is the user's middle initial\\ Which team has 41 points?}} \\
        \cmidrule(lr){1-2}
        Subtotal & 83.8 & \multicolumn{2}{l}{}  \\
    \bottomrule
    \end{tabular}
    \caption{Top question category distribution ($\ge 2.0\%$) and examples (See Appendix \ref{appendix:dataset_analysis} for remaining categories).}
    \label{tab:top_q_categories}
\end{table*}

Among the~86k collected questions,~47.5k are unique in term of SQuAD EM. 
Common questions often recur across screenshots (e.g., ``Which option is selected?'' or ``What is the email address?''). 
Screenshots with more information typically generate more questions. 
The histogram of questions per screenshot follows a mild exponential decay (Figure~\ref{fig:q_num}).
We further categorize questions using regular expressions, providing a preliminary overview in Table~\ref{tab:top_q_categories}.
It is worth noting that the category distribution is implicitly influenced by RICO's crawling process.
For example, the crawling process commonly includes a step at the login page, resulting in a higher percentage of questions about app names, email addresses, and login permissions.



\subsection{Answer Analysis}

We analyze the answer annotations in two aspects:~1) How often multiple bounding boxes are required for answers, indicating task complexity, and ~2) How often VH is sufficient for bounding box annotation, indicating the reliability of VHs.

Figure~\ref{fig:a_num} illustrates the histogram of bounding boxes per answer.
About~84\% of answers contain a single bounding box, with 51\% utilizing VH leaf nodes and 49\% using manually drawn boxes. 
When all answers are considered, 51\% rely solely on VH leaf nodes, 48\% on manual boxes, and 0.8\% on a combination. 
Combinations typically occur for multi-part answers involving diverse UI elements, scattered components (e.g., dates), or measurements with units.
Despite VHs being commonly used in prior screen tasks~\cite{burns2022motifvln, liWidgetCaptioningGenerating2020}, the near-equal preference for manual bounding boxes reflect VH limitations: VHs cannot capture UI elements in WebViews and Canvas, and are inconsistent for certain screen designs.
This justifies our decision to use screenshot as input
in 
ScreenQA.



\section{Experiments and Baselines}
\label{sec:baselines}

We conducted three sets of experiments: zero-shot, fine-tuning, and transfer learning, each of which is explained below.


\begin{table}[t]
    \centering
    \small
    \begin{tabular}{lcc} %
        \toprule
        & SQuAD-EM & SQuAD-F1 \\
        \midrule
        Fuyu-8B & 39.5 & 47.3  \\
        Gemini 1.5 Flash & 80.6 & 86.4 \\
        Gemini 1.5 Pro & \textbf{81.4} & \textbf{87.2} \\
        GPT-4o & 77.8 & 86.6 \\
        \bottomrule
    \end{tabular}
    \caption{
        Zero-shot public models evaluation on SQA-S.
        Bold is best performance.
    }
    \label{tab:zero_shot_sqas_results}
\end{table}

\begin{table*}[t]
    \centering
    \small
    \begin{threeparttable}
    \begin{tabular}{lcccccccccc} %
        \toprule
        \multirow{2}{*}{\raisebox{-3pt}{Model}} & \multicolumn{2}{c}{SQA-S} & \multicolumn{3}{c}{SQA-L} & \multicolumn{2}{c}{SQA-UIC} & \multicolumn{3}{c}{SQA-UIC-BB} \\
        \cmidrule(lr){2-3}
        \cmidrule(lr){4-6}
        \cmidrule(lr){7-8}
        \cmidrule(lr){9-11}
        & EM & F1 & R-1 & R-2 & R-L & EM & F1 & BBOX-F1 & EM & F1 \\
        \midrule
        ScreenAI 670M & 51.2 & 60.6 & 77.3 & 68.4 & 76.7 & 47.8 & 49.4 & 62.7 & 41.1 & 42.6 \\
        PaliGemma 3B 224 & 77.5 & 83.9 & 88.2 & 81.5 & 87.4 & 74.8 & 76.7 & 84.9 & 67.5 & 69.6 \\
        PaliGemma 3B 448 & 88.3 & 92.2 & 91.1 & 85.5 & 90.3 & 86.0 & 87.7 & 89.4 & 79.1 & 81.6 \\
        PaliGemma 3B 896 & 89.4 & 93.2 & 90.9 & 85.3 & 90.1 & 86.1 & 87.8 & 88.8 & 78.8 & 81.2 \\
        ScreenAI 5B & \textbf{90.7} & 94.6 & \textbf{92.6} & \textbf{87.4} & \textbf{91.9} & 87.0 & 88.7 & \textbf{94.2} & \textbf{84.0} & \textbf{85.7} \\
        Gemini 1.5 Flash & 90.5 & \textbf{94.9} & 92.4 & 86.2 & 91.7 & \textbf{88.2} & \textbf{89.7} & 92.4 & 83.9 & \textbf{85.7} \\
        \bottomrule
    \end{tabular}
    \end{threeparttable}
    \caption{Model performance for ScreenQA tasks after fine-tuning. Bold is best performance.}
    \label{tab:finetuned_results}
\end{table*}

\begin{table*}[ht]
    \centering
    {
    \footnotesize
    \begin{tabular}{lcccccccc} 
        \toprule
        Experiment | Task & \multicolumn{4}{c}{\textbf{CDL} | SQA-S} & \multicolumn{2}{c}{\textbf{CDL} | DocVQA} & \multicolumn{2}{c}{\textbf{TL} | VisualWebBench-WebQA} \\
        \cmidrule(lr){2-5}
        \cmidrule(lr){6-7}
        \cmidrule(lr){8-9}
        Fine-Tuned on & \multicolumn{2}{c}{SQA-S} & \multicolumn{2}{c}{DocVQA} & SQA-S & DocVQA & DocVQA & SQA-S + DocVQA \\
        \cmidrule(lr){2-3}
        \cmidrule(lr){4-5}
        \cmidrule(lr){6-6}
        \cmidrule(lr){7-7}
        \cmidrule(lr){8-8}
        \cmidrule(lr){9-9}
        Model $\backslash$ Metrics & EM & F1 & EM & F1 & ANLS & ANLS & F1 & F1\\
        \midrule
        PaliGemma 3B 224 & 77.5 & 83.9 &52.6 & 60.5 & 27.5 & 43.7 & 19.31 & \textbf{21.51} \\
        PaliGemma 3B 448 & 88.3 & 92.2 & 66.2 & 72.9 & 55.1 & 78.0 & 48.33 & \textbf{49.11} \\
        PaliGemma 3B 896 & \underline{89.4} & 93.2 & \underline{63.6} & 70.4 & \underline{62.0} & \underline{84.8} & 57.07 & \textbf{58.69} \\
        \bottomrule
    \end{tabular}
    }
    \caption{
        Cross-domain learning (CDL) and transfer learning (TL) experiments via PaliGemma model fine-tuning.
        For CDL, models were fine-tuned on either SQA-S or DocVQA and evaluated on the other.
        In-domain and cross-domain model scores exceed~60 for the~896$\times$869 model and differ by~20$\sim$25 points across the board (underlined).
        TL experiments revealed positive transfer when additionally using SQA-S for the WebQA evaluation (bold).
    }
    \label{tab:cdl_tl}
\end{table*}

\begin{table*}[h!t]
    \centering
    \small
    \begin{tabular}{lcccccccccc} %
        \toprule
        \multirow{2}{*}{\raisebox{-3pt}{Evaluation setup}} & \multicolumn{2}{c}{SQA-S} & \multicolumn{3}{c}{SQA-L} & \multicolumn{2}{c}{SQA-UIC} & \multicolumn{3}{c}{SQA-UIC-BB} \\
        \cmidrule(lr){2-3}
        \cmidrule(lr){4-6}
        \cmidrule(lr){7-8}
        \cmidrule(lr){9-11}
        & EM & F1 & R-1 & R-2 & R-L & EM & F1 & BBOX-F1 & EM & F1 \\
        \midrule
        Zero-Shot, Text-Only & 64.4 & 72.5 & 78.2 & 67.8 & 75.6 & 38.6 & 41.2 & 30.1 & 49.7 & 28.5 \\
        Zero-Shot & 80.6 & 86.4 & 83.8 & 70.8 & 79.3 & 62.4 & 66.8 & 26.2 & 33.3 & 24.0 \\
        Fine-Tuned & 90.5 & 94.9 & 92.4 & 86.2 & 91.7 & 88.2 & 89.7& 85.7 & 92.4 & 83.9 \\
        \bottomrule
    \end{tabular}
    \caption{
        Gemini 1.5 Flash performance against ScreenQA for various evaluation setups. Note that for SQA-UIC-BB, the Zero-Shot, Text-Only case (the 1st row) outperforms its multimodal counterpart (the 2nd row) because the bounding box coordinates are given as text input: the model just needs to select the right UI elements with the given coordinates, which makes bounding box prediction very precise when they are correct, hence, an easier task.
    }
    \label{tab:gemini_1_5_flash_evals}
\end{table*}


\subsection{Zero-Shot Experiments}

We evaluated 
Fuyu-8B~\cite{fuyu-8b}, Gemini 1.5 Flash, Gemini 1.5 Pro \cite{team2023gemini}, and GPT-4o~\cite{openai2024gpt4} on SQA-S in a zero-shot setting. 
Results are summarized in Table~\ref{tab:zero_shot_sqas_results}. 
We prompted each model using an instruction followed by a question.
For the Fuyu-8b model, we used the instruction that model's authors recommended in their examples.
For GPT-4o, we iterated on the prompt design using a set of 10 examples from the ScreenQA validation split and then reused the prompt for Gemini 1.5 models.
See Appendix~\ref{appendix:zero_shot_evaluation_details} for the detailed prompts.
Despite the prompt being originally designed for GPT-4o, Gemini 1.5 Pro outperforms GPT-4o in this particular setup.



\subsection{Fine-Tuning Experiments}

We fine-tuned three series of models spanning open-source, domain specific, and general~purpose proprietary models on the ScreenQA tasks as described below.

\paragraph{PaliGemma 3B}
We used the pre-trained PaliGemma 3B model checkpoints~\cite{beyer2024paligemma} with three input resolutions, 224$\times$224, 448$\times$448 and 896$\times$896,
and fine-tuned for~10 epochs with a learning rate $1.0\times10^{-5}$ using the Adam optimizer and a cosine decay schedule.
Both vision and language backbones were trained during fine-tuning.

\paragraph{ScreenAI 670M and 5B}
This VLM specializes in UI and infographics understanding~\cite{baechler2024screenai}. We used a dynamic $812\times812$ input resolution and 
fine-tuned this model until convergence, using a learning rate of $1.0\times10^{-3}$.
Only the language backbone was trained during fine-tuning.

\paragraph{Gemini 1.5 Flash}
We also fine-tuned Gemini 1.5 Flash~\cite{team2023gemini} model\footnote{\scriptsize{\href{https://cloud.google.com/vertex-ai/generative-ai/docs/models/gemini-use-supervised-tuning}{\texttt{https://cloud.google.com/vertex-ai/generative-ai}} \\
\href{https://cloud.google.com/vertex-ai/generative-ai/docs/models/gemini-use-supervised-tuning}{\texttt{/docs/models/gemini-use-supervised-tuning}}}}
on the downstream tasks for $\sim7$ epochs with a learning rate of $1.0\times10^{-4}$ for SQA-S and SQA-L and $1.0\times10^{-6}$ for SQA-UIC and SQA-UIC-BB.
Only the language backbone was trained during fine-tuning.

Note that we did not test PaliGemma and ScreenAI models' zero-shot performance as they are not instruction-tuned, therefore, not suitable for evaluation in a zero-shot setting.

We use the metrics introduced in Section~\ref{sec:tasks_metrics} to measure the performance of models along various dimensions: i) Extracting relevant information for answering a question in the SQA-S and SQA-UIC tasks, ii) Ability to provide fluent answers in the SQA-L task, and iii) Identifying relevant UI elements through their bounding box coordinates in the SQA-UIC-BB task.

We present the results in Table~\ref{tab:finetuned_results}.
We observe a slightly higher performance for the ScreenAI~5B model compared to PaliGemma~3B, and attribute it to its larger model capacity and specialized pre-training mixture that includes a rich variety of UI elements. The ScreenAI model also performs on par with the Gemini 1.5 Flash model in a fine-tuned setting. We also notice a larger difference between the ScreenAI model and other approaches in the SQA-UIC-BB tasks involving bounding box prediction.
PaliGemma performance is however very competitive, and by fine-tuning both language and vision backbones we enable better use of the entire model capacity.

In Appendix~\ref{appendix:observed_prediction_errors} we present a few examples of the different errors made by the models and a categorization of these errors.
 
\subsection{Cross-Domain and Transfer Learning}

We conducted cross-domain learning (CDL) and transfer learning (TL) by fine-tuning PaliGemma models (Table~\ref{tab:cdl_tl}).
For CDL, we fine-tuned the models using either SQA-S or DocVQA~\cite{mathew2021docvqa} and evaluated on the other.
These experiments aim to characterize the differences between ScreenQA and other VQA datasets from related domains like document images by investigating the performance difference by training on related datasets vs. training on the corresponding training split.
We observe that the use of in-domain training data results in significant improvements for both SQA-S and DocVQA datasets.
However, we do note that for the SQA-S task, using training data from the related domain of document images results in performance gain over the zero-shot setting with a larger Fuyu-8B model (Table~\ref{tab:zero_shot_sqas_results}), showing that they are related but distinct domains. 

For transfer learning, summarized in Table~\ref{tab:cdl_tl}, we used VisualWebBench-WebQA~\cite{liuVisualWebBenchHowFar2024}, a recently released QA task on web~screenshots, as the target evaluation dataset.
We fine-tuned the models using DocVQA alone and using both DocVQA and ScreenQA, demonstrating positive transfer when incorporating the ScreenQA dataset.

Additionally, we tested the hypothesis that training models to predict bounding boxes would enhance their performance on tasks not explicitly involving bounding boxes.
For example, fine-tuning models on both SQA-S and SQA-UIC-BB tasks and evaluating performance on SQA-S task.
However, in our experiments with smaller models such transfer benefits did not materialize.
One possible explanation could be that this is due to increased complexity caused by introducing an additional task.
While this may seem like a limitation, it also reflects the standalone challenges each task presents, reaffirming the value of our dataset.

\subsection{Importance of Multimodality and Fine-Tuning}

We used Gemini 1.5 Flash~\cite{team2023gemini} model and SQA-S and SQA-L tasks to demonstrate the complexity of the screen data for screen understanding.
Experiments in Table~\ref{tab:gemini_1_5_flash_evals} show that using text representation of the screen information (OCR) as model input produces significantly worse results than using multimodal (image \& text) setup.
Noticeable difference in performance between zero-shot and fine-tuned setups suggests that screen understanding domain presents its own challenges on top of generic image understanding.

\section{Conclusion}
\label{sec:conclusion}

We introduced ScreenQA, a rich dataset that enables training and evaluating models on question-answering tasks on screen content.
We described the annotation process, statistics of the collected dataset, which contains 85,984 question-answer pairs.
In addition to answers, our dataset contains extensive annotations of the UI elements, enabling the ability to train or probe models for their holistic understanding of the screen, a necessary capability for high-level reasoning and automation using UI interfaces.
Compared to other vision-language tasks, such as document understanding or visual-question answering, the four constructed tasks on the ScreenQA dataset pose unique challenges: rich in text, diverse in mobile applications, blended with icons and symbols.
The tasks not only evaluate content quality, but also UI element identification quality.
Furthermore, we conducted a diverse set of experiments, including zero-shot, fine-tuned, and transfer learning, on a series of open-weight and proprietary models to assess the capacity of our ScreenQA dataset.
We encourage the community to tackle screen content understanding challenges present in our benchmark, fostering new technologies and user experiences.

\section{Limitations}\label{sec:limitations}

We acknowledge limitations of our work.

\paragraph{Language} Released data is only in English and further work would be necessary to create a multilingual variant of our dataset. This is further amplified by the need of screenshots with phones configured in the different locales.

\paragraph{Rich Layouts} Our data and annotation process focuses on rather static content, facilitating us to focus on information extraction from canonical layouts. There are a number of challenges our dataset does not capture, specifically rich layouts coming from gaming or other highly interacting applications that may be present on a user's device. Screenshots that contain natural images and videos are also missing.

\paragraph{Reasoning} The type of challenges our dataset focuses on are a lot more related to information look-up and rewriting. We therefore capture challenges grounded in UI content and composition understanding, rather than arithmetic or other forms of complex compositional reasoning challenges. See Section~\ref{sec:question_annotation} for details.

\paragraph{Multi-Image} An example in our dataset always consists of a single screenshot. Several challenges that come with multi-image understanding such as a trace of screenshots are therefore not covered by our tasks in the dataset. Similarly, no UI animations are included or scrolling actions.

\paragraph{Platform} ScreenQA contains screenshots only from the Android ecosystem and the phone form factor.
As all of the composed questions are focused on the main content area without involving any gestures and actions, we do not anticipate major performance differences across ecosystems. In addition, our experiments show positive transfer for VisaulWebBench-WebQA when additionally using ScreenQA (Table~\ref{tab:cdl_tl}), indicating the efficacy of ScreenQA for other form factors.

\section{Ethical Considerations}

\paragraph{Information Retrieval Only}
ScreenQA focuses on information retrieval of screen contents, with the primary intent of improving screen content understanding. 
ScreenQA dataset does not involve any materials related to decision-making for or against users, nor does it execute actions on behalf of users, avoiding potential harm.

\paragraph{Privacy}
The technologies fostered by ScreenQA, when used on a mobile device, require access to information at a level comparable to that of a standard accessibility application.
Furthermore, the promising results from fine-tuned PaliGemma models (Table~\ref{tab:finetuned_results}) suggest that the technologies enabled by ScreenQA can be effectively hosted on mobile devices, significantly limiting potential privacy concerns.

\paragraph{App Category Fairness}
ScreenQA contains questions across~27 categories of apps (Figure~\ref{fig:q_num}), which is the most diverse set at the time of writing (Table~\ref{tab:rico_comparison}).
Also, our fine-tuned PaliGemma~896 model exhibits consistent SQuAD-F1 performance across these app categories (Figure~\ref{fig:sqa_s_metric_per_category}), demonstrating that ScreenQA has a fair support across app purposes.


\clearpage

\appendix

\section{Data Annotation Details}
\label{appendix:annotation_ui}

\subsection{VH Out-of-Sync Rules} \label{appendix:prefiltering_rules}

\begin{figure*}[ht]
    \centering
    \begin{minipage}[t]{\textwidth}
        \centering
        \begin{subfigure}[t]{0.27\textwidth}
            \centering
            \includegraphics[width=\textwidth]{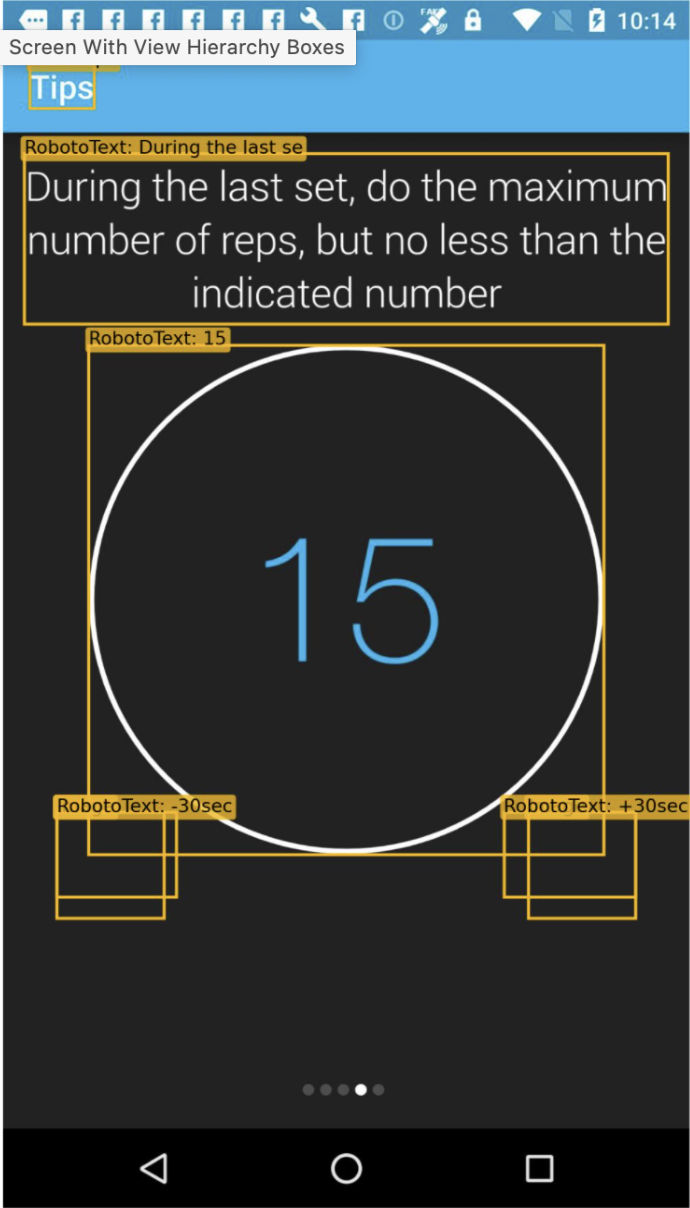}
            \caption{VH with occluded elements.}
            \label{fig:insync-occlusion}
        \end{subfigure}\hfill
        \begin{subfigure}[t]{0.27\textwidth}
            \centering
            \includegraphics[width=\textwidth]{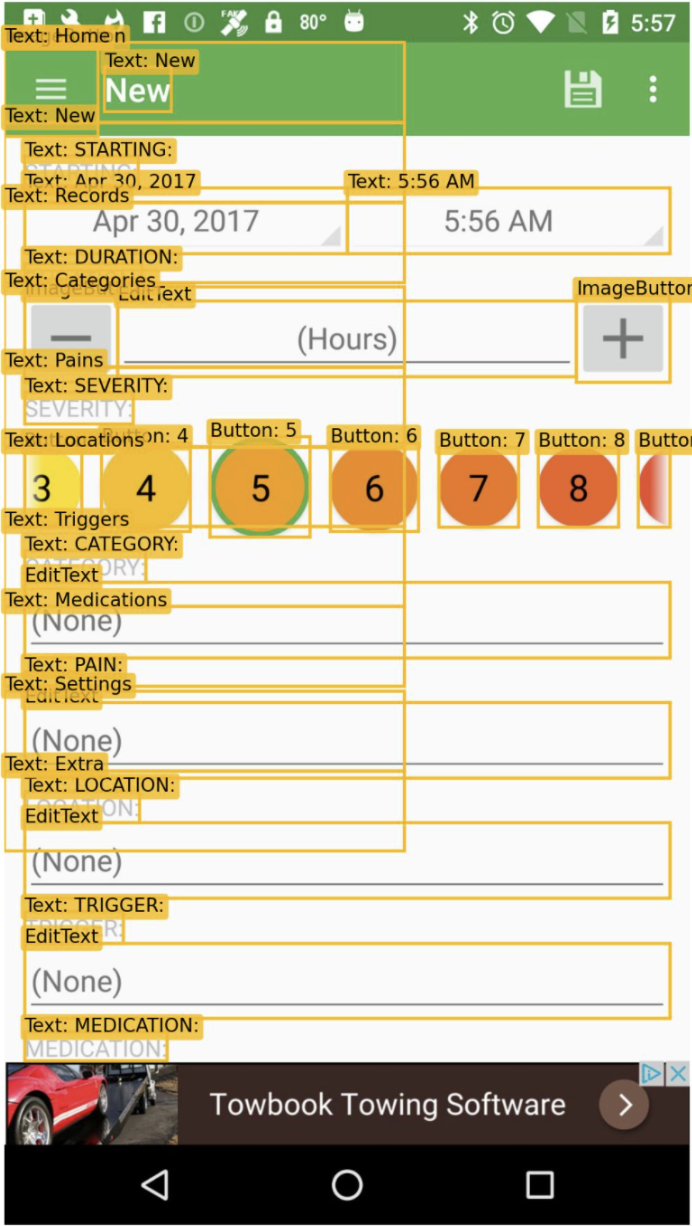}
            \caption{Ghosting VH from menu.}
            \label{fig:insync-ghosting}
        \end{subfigure}  \hfill
        \begin{subfigure}[t]{0.27\textwidth}
            \centering
            \includegraphics[width=\textwidth]{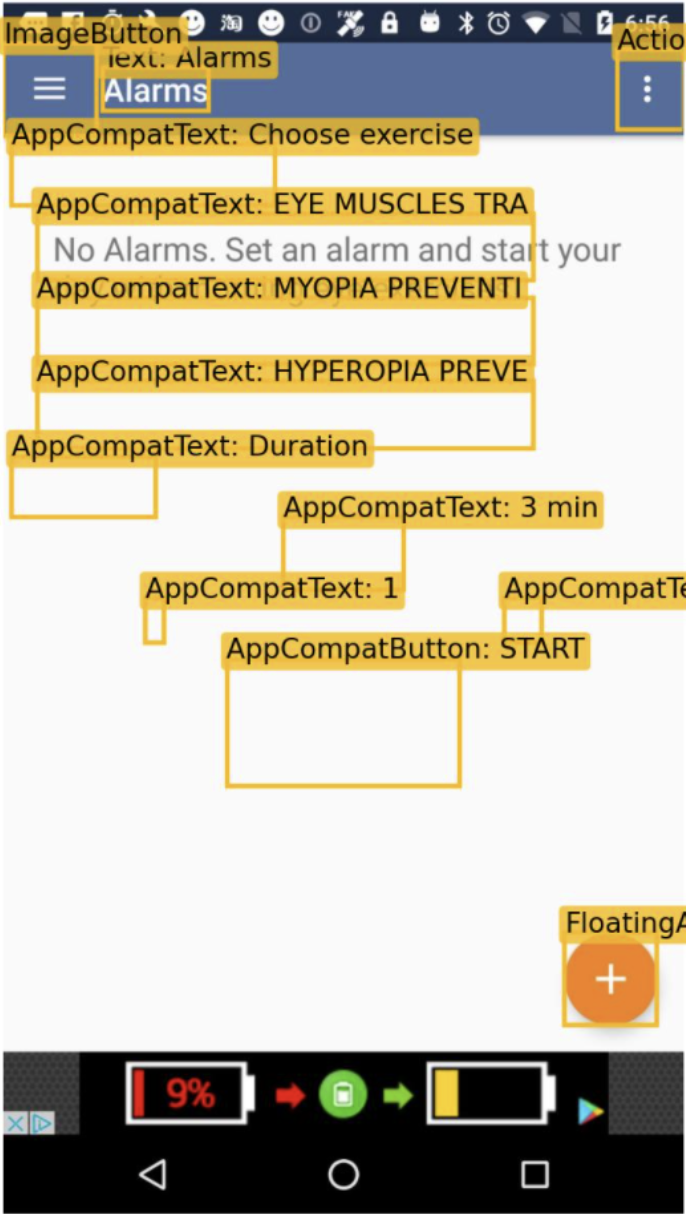}
            \captionsetup{belowskip=10pt}
            \caption{Out-of-sync VH for main content.}
            \label{fig:outofsync}
    \end{subfigure}\hfill
    \caption{
        View hierarchies (VHs) are overlaid on the screenshots with class names and the first few characters printed to assist annotators to determine whether the VHs for the main contents are in sync.
    }
    \label{fig:vh_sync}
    \end{minipage}
\end{figure*}

We asked annotators to mark screenshots which are 1) from non-English apps and 2) not synchronized with VHs, as described in Section~\ref{sec:pre-filtering}.
The out-of-sync symptoms are oftentimes harder to distinguished than expected.
UI elements may be occluded (Figure~\ref{fig:insync-occlusion}) and ``ghosting'' VHs appear in addition to the in-sync VHs because of UI animation effects such as menus sliding/popping in and out (Figure~\ref{fig:insync-ghosting}).
These two cases are deemed to be acceptable in-sync scenarios as a user or an annotator can still select the correct bounding boxes from VH.
However, the example in Figure~\ref{fig:outofsync} is an out-of-sync case, in which the text ``No Alarms. Set an alarm and start your ...'' is not supported by a VH bounding box.
All the other bounding boxes that appear on the screen are all irrelevant to the current main content, hence, determined as an out-of-sync case.
We trained our annotators to distinguish these cases and only mark the third case to remove.

\subsection{Question Annotation UI}
\label{appendix:question_annotation_ui}

\begin{figure*}[t]
    \centering
    \begin{subfigure}[t]{0.8\textwidth}
        \centering
        \includegraphics[width=\textwidth]{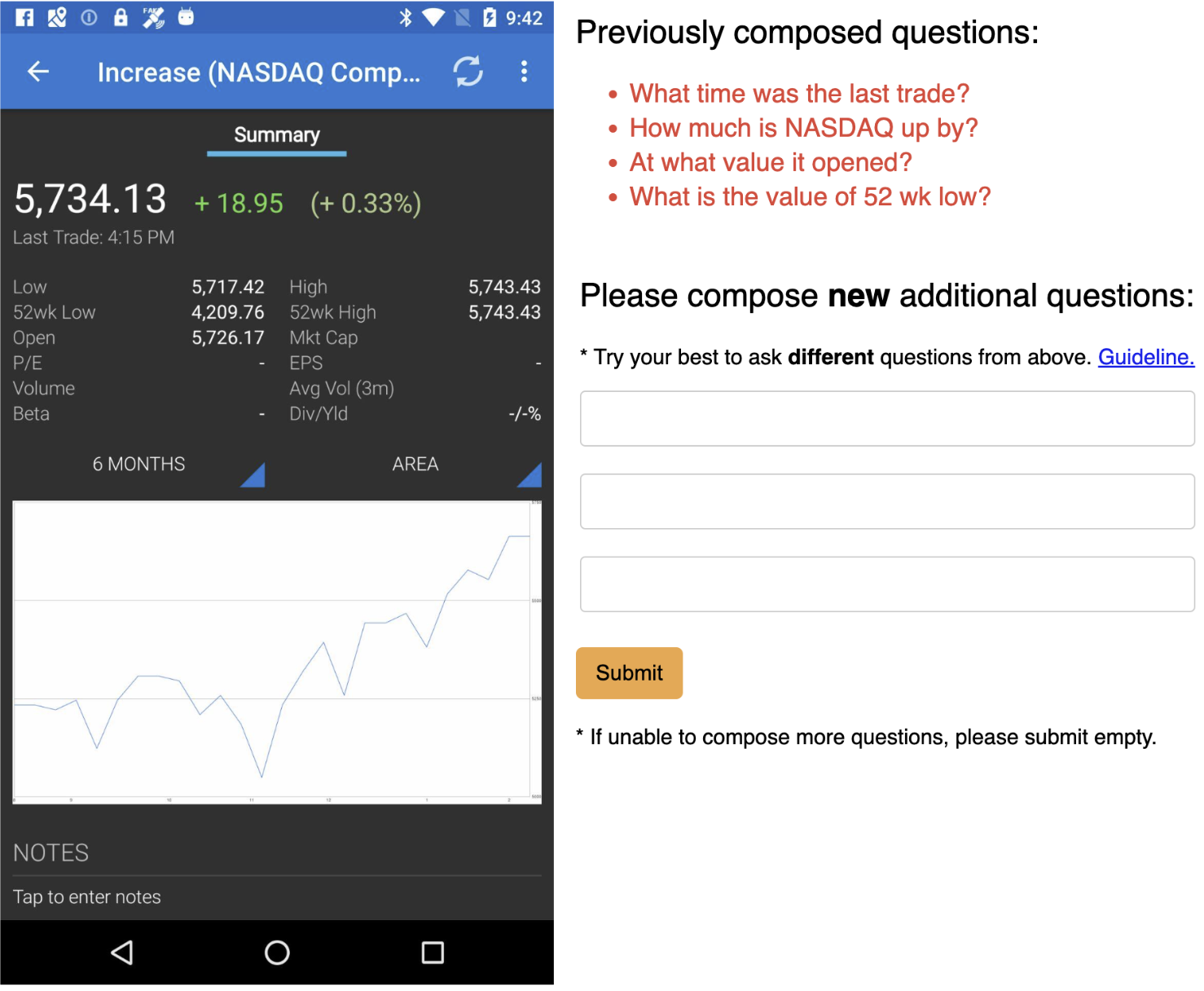}
        \caption{Question annotation UI}
        \label{fig:question_annotation_ui}
    \end{subfigure}\hfill
    \begin{subfigure}[t]{0.8\textwidth}
        \centering
        \includegraphics[width=\textwidth]{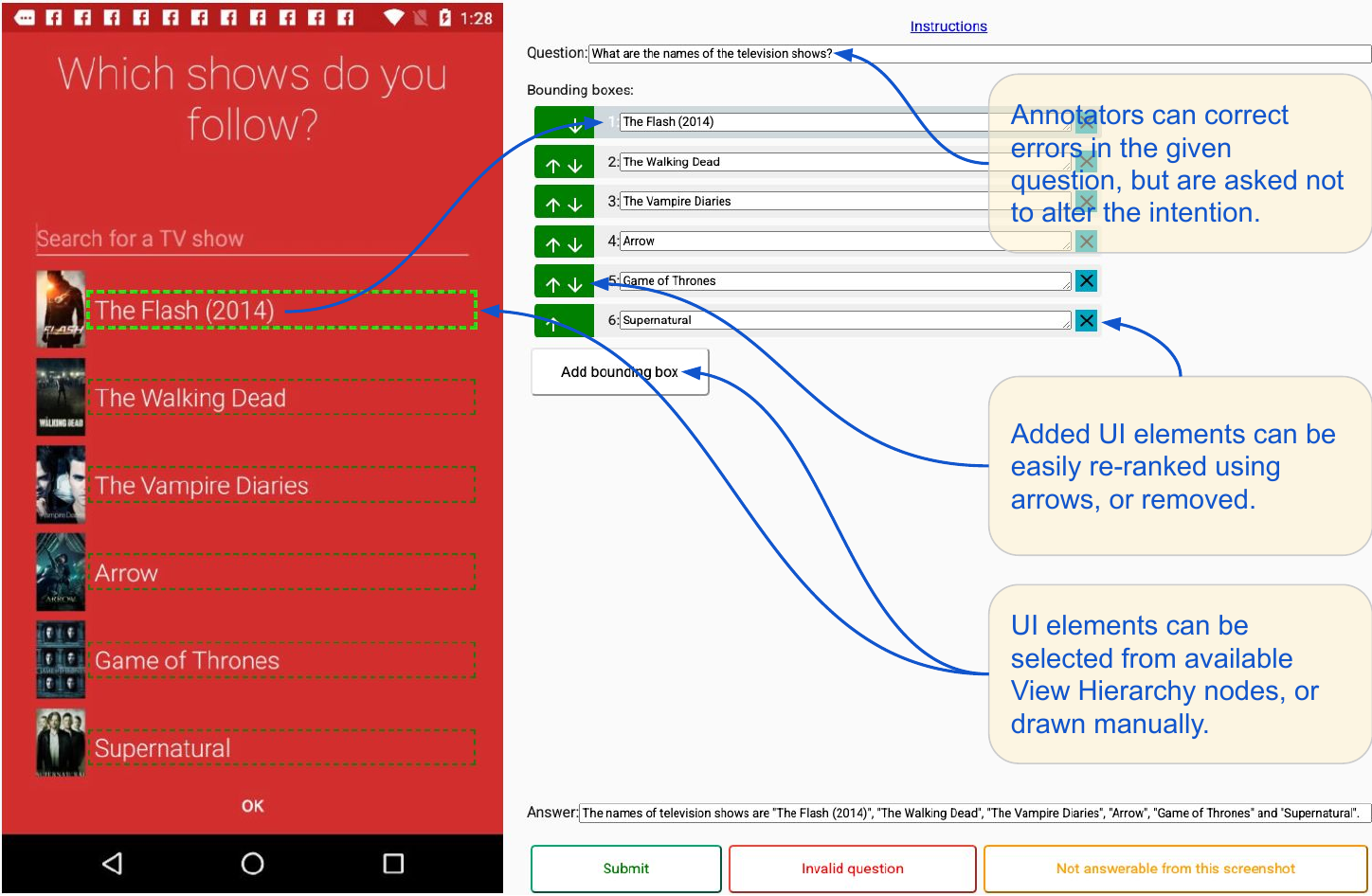}
        \caption{Answer annotation UI}
        \label{fig:answer_annotation_ui}
    \end{subfigure}\hfill
    \caption{Annotation interfaces.}
    \label{fig:annotation_ui}
\end{figure*}

The question annotation interface is shown in Figure~\ref{fig:question_annotation_ui}.
Question annotation was performed in a sequential manner by multiple annotators. An annotator can see all previous questions to diversify question framing and avoid duplication.
We also used the same sequential process to provide more feedback and training to the annotators for quality improvement.

\subsection{Answer Annotation UI}
\label{appendix:answer_annotation_ui}


The answer annotation interface is shown in Figure~\ref{fig:answer_annotation_ui}.
Answer annotators were tasked to determine if the question is valid and if the question is answerable from the screen context.
If both are positive, the annotators need to answer the questions by 1)~selecting or drawing the bounding boxes of UI elements, 2)~filling the text for each selected/drawn bounding box on the right, 3)~ranking them appropriately, 4)~providing the full-sentence answer to the question.
The annotators were also tasked to review and make necessary corrections if the question has grammatical errors or typos.

\subsection{Annotation quality control}
\label{appendix:annotation_quality_control}

We leveraged a rating platform for collecting human annotations to ensure that raters go through a training phase, and proceed to collecting annotations only after reaching 90\% quality bar.
Furthermore, output quality was monitored throughout the annotation process, consistently reaching more than 95\%.

\subsection{Annotation post-processing}
\label{appendix:annotation_post_processing}

Aside from removing answer annotations marked as ``invalid question'' and questions with no remaining answer annotation, post-processing of the records in the train split included applying question modification, if there was one provided during answer annotation.

The same could not be easily done for validation and test splits, as those contained 3 answer annotations per question, and corresponding question modifications could vary.
The inter-annotator agreement was therefore estimated.
The 2 answer annotations were considered to be in agreement if any of the following conditions apply:
\begin{itemize}
    \item The questions (after modifications if some were provided) are the same.
    \item The full-sentence answers are the same.
    \item The lists of bounding boxes that constitute the answer contain the same elements, ignoring permutations.
    \item Both answer annotations are in agreement with the same other answer annotation.
\end{itemize}
Here two bounding boxes correspond to the same element if they correspond to the same VH node, or if they intersect and have the same textual content.
As a result, 97.9\% of the questions in validation and test splits had full agreement of the answer annotations, 1.9\% had partial agreement (2 out of 3), and 0.2\% had no agreement.

We then removed the disagreeing answer annotations from questions with partial agreement, and questions with no agreement.
And applied question modification with the biggest consensus (2-3 out of 3), or chose the latest if all modifications are different.

\subsection{Short Answer Generation Prompts}
\label{appendix:screenqa_short_prompts}

We describe below the prompts used in PaLM 2~\citep{anil2023palm} to generate short answers for ScreenQA, as described in Section~\ref{sec:short_answers_generation}.
ScreenQA dataset annotations (question, list of text of UI elements,  and full-sentence answer) were used as input.
To improve the prompt quality, we used an identical prompt template with two sets of few-shot examples for answers with 1) one or 2) more than one UI elements involved, each set of which is used and inserted into the template for the example with answers that utilize the corresponding number of UI elements.
The prompt template is give below:



\lstset{
  basicstyle=\scriptsize\ttfamily,
  breakindent=0pt,
  breaklines=true,
  literate=
    {°}{{$\SI{}{\degree}$}}1
    {®}{{\textsuperscript{\textregistered}}}1
    {™}{{\texttrademark}}1
    {⇆}{{$\rightleftarrows$ }}2
}

\begin{lstlisting}
List various ways to rephrase the answer. The answer should be as short as possible, without extra words from the question. Use all provided elements in each answer. Provide the output in square brackets.

{examples}

Now is your turn.
Question: {question}
Answer elements: {list of text of UI elements}
Full answer: {full-sentence answer}
Rephrases:
\end{lstlisting}

\noindent An example of single-UI-element answer is as below:
\begin{lstlisting}
Here is another example:
Question: 'What is the gender?'
Answer elements: ['Male']
Full answer: 'The gender is male.'
Rephrases: ['male']
\end{lstlisting}

\noindent  An example of multiple-UI-element answer is as below:
\begin{lstlisting}
Here is another example:
Question: 'What is the name?'
Answer elements: ['Jon', 'Brown']
Full answer: 'The name is Jon Brown.'
Rephrases: ['Jon Brown']
\end{lstlisting}

\section{Data Examples}
\label{appendix:data_examples}

Table~\ref{tab:examples1} presents a few examples from the ScreenQA dataset.
Each example contains 
\begin{itemize}[nolistsep, leftmargin=*]
\item A screenshot
\item A question
\item Multiple annotations of lists of UI elements relevant to the question, each annotation of which is completed by an annotator
\item Multiple annotations of long full-sentence answers corresponding to the UI element annotations
\item Multiple short answers, generated by the procedure outlined in Section~\ref{sec:short_answers_generation}
\end{itemize}
Note that the bounding boxes of selected UI elements are highlighted on the screenshot for the illustrated purposes of the corresponding annotated bounding boxes, but they are not present in the corresponding image from RICO~\cite{dekaRicoMobileApp2017} or during our data annotation process outlined in Section~\ref{sec:data_annotation}.

\clearpage
\onecolumn
\begin{longtable}{ll}
\\
    \toprule
    \raisebox{-42mm}{\includegraphics[height=75mm]{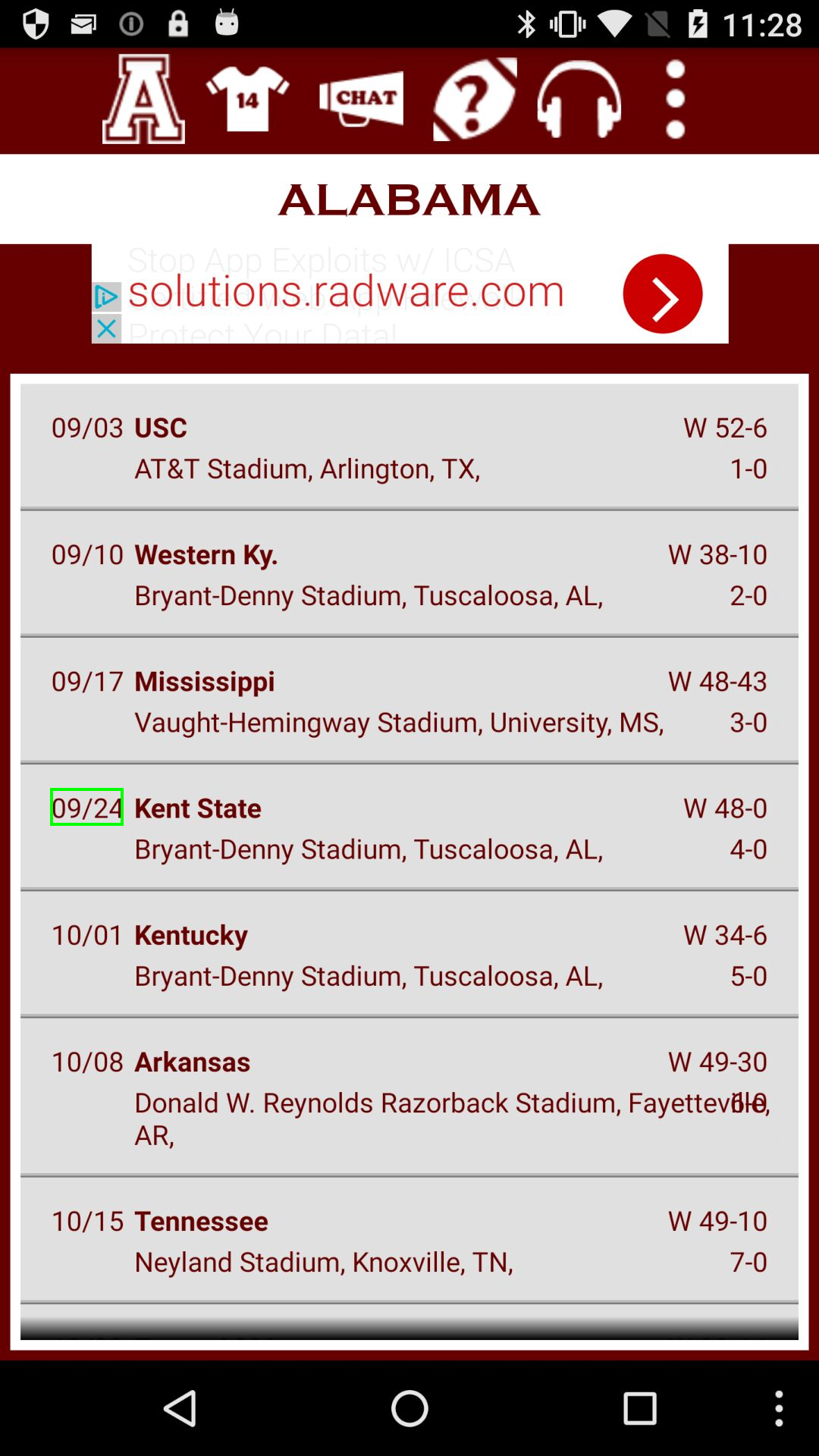}}
    & 
    \begin{minipage}{0.65\textwidth}
        \begin{tabular}{ll}
            Question & When was the match held at Kent State?
            \\ \\
            UI elements & 
                \makecell[tl]{ 
                    •\; {[09/24]}
                    •\; {[09/24]}
                } 
            \\ \\
            Full answers &
                \makecell[tl]{ 
                    •\; The match was held at Kent State on \\ \quad September 24. \\
                    •\; The match was held on September 24.
                }
            \\ \\
            \makecell[tl]{Generated\\ Short Ans.} &
                \makecell[tl]{
                    •\; 09/24 \\
                    •\; September 24 \\
                    •\; September 24th \\
                    •\; 9/24
                } 
            \\ \\
        \end{tabular}
    \end{minipage} \\    
    \midrule
    \raisebox{-35mm}{\includegraphics[height=75mm]{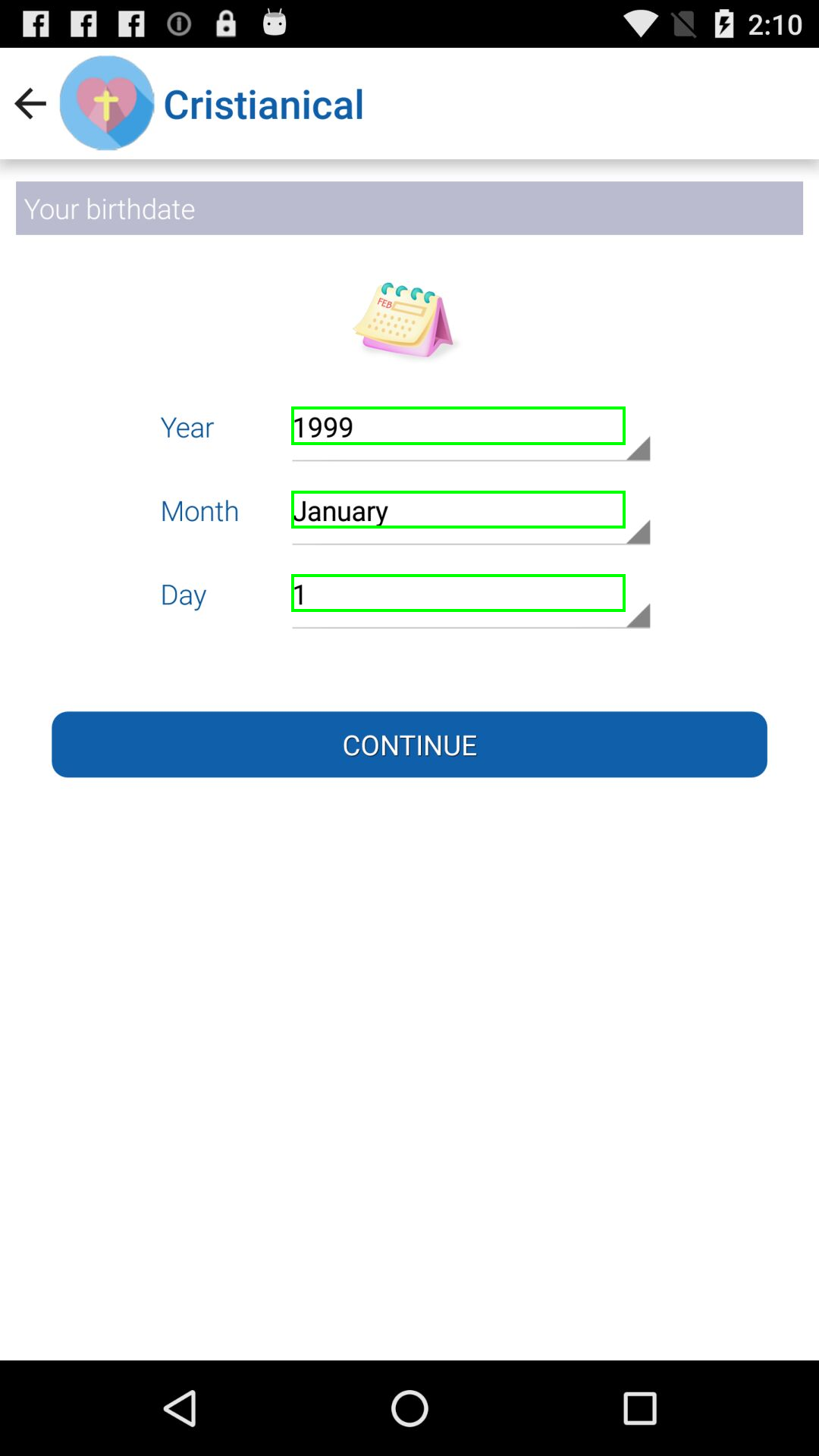}}
    & 
    \begin{minipage}{0.65\textwidth}
        \begin{tabular}{ll}
            Question & What is the birth date of the user?
            \\ \\
            UI Elem. & 
                \makecell[tl]{ 
                    •\; {[1999]}, {[January]}, {[1]} \\
                    •\; {[January]}, {[1]}, {[1999]} \\
                    •\; {[1]}, {[January]}, {[1999]}
                } 
            \\ \\
            Long Ans. &
                \makecell[tl]{ 
                    •\; The birth date of the user is January 1, 1999. \\
                    •\; The user's birth date is January 1, 1999. \\
                    •\; The birth date is January 1, 1999.
                }
            \\ \\
            \makecell[tl]{Generated\\ Short Ans.} &
                \makecell[tl]{
                    •\; 1/1/1999 \\
                    •\; January 1, 1999 \\
                    •\; 1 January 1999 \\
                    •\; 1 January, 1999 \\
                    •\; January 1st, 1999
                } 
            \\ \\
        \end{tabular}
    \end{minipage} \\
    \midrule
    \raisebox{-50mm}{\includegraphics[height=75mm]{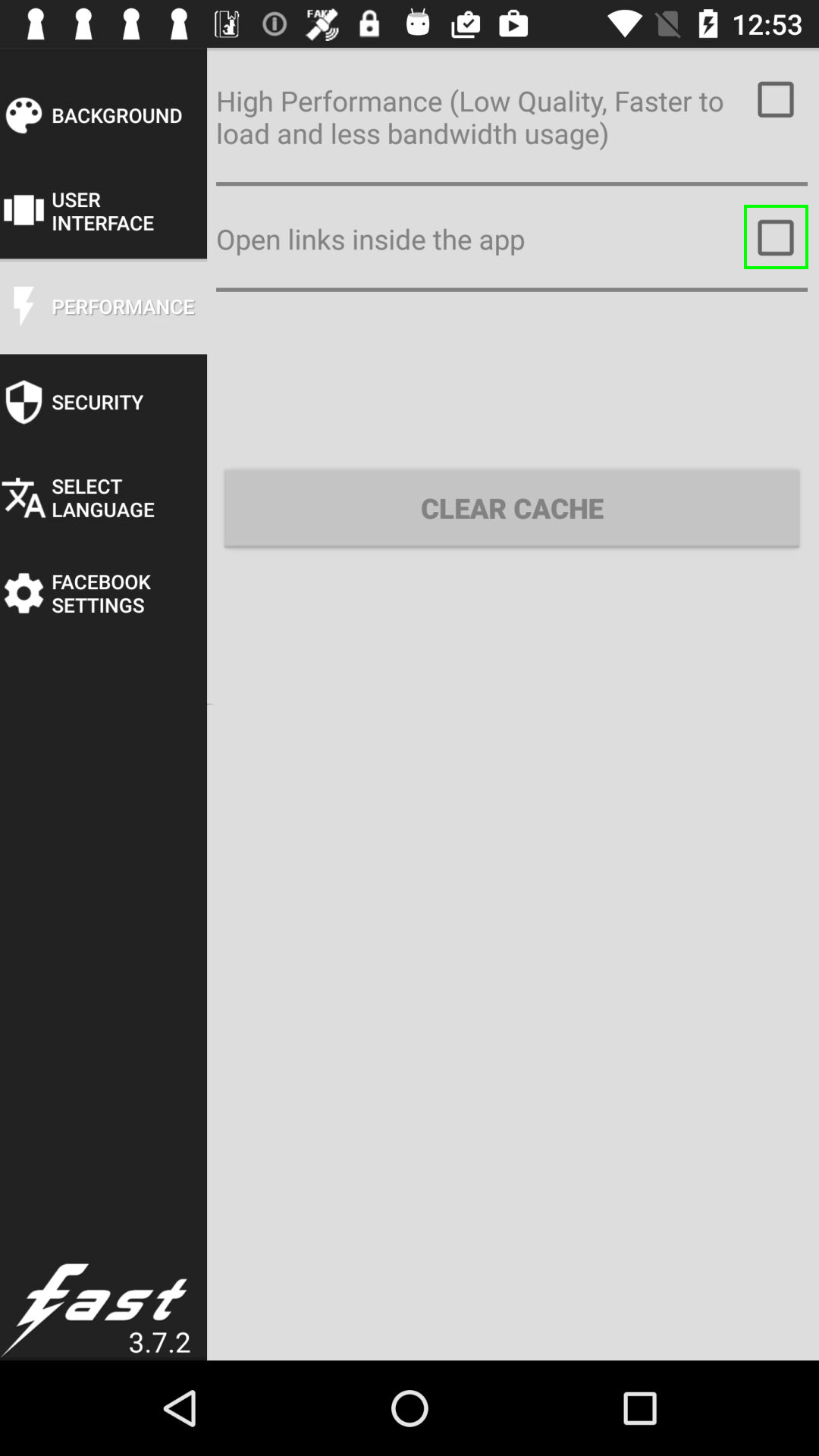}}
    & 
    \begin{minipage}{0.65\textwidth}
        \begin{tabular}{ll}
            Question & What is the status of "Open links inside the app"?
            \\ \\
            UI Elem. & 
                \makecell[tl]{ 
                    •\; {[off]}
                    •\; {[off]}
                } 
            \\ \\
            Long Ans. &
                \makecell[tl]{ 
                    •\; The status of "Open links inside the app" is "off". \\
                    •\; The status is "off".
                }
            \\ \\
            \makecell[tl]{Generated\\ Short Ans.} &
                \makecell[tl]{
                    •\; off \\
                    •\; disabled
                } 
            \\ \\
        \end{tabular}
    \end{minipage} \\
    \midrule
    \raisebox{-50mm}{
        \includegraphics[height=75mm]{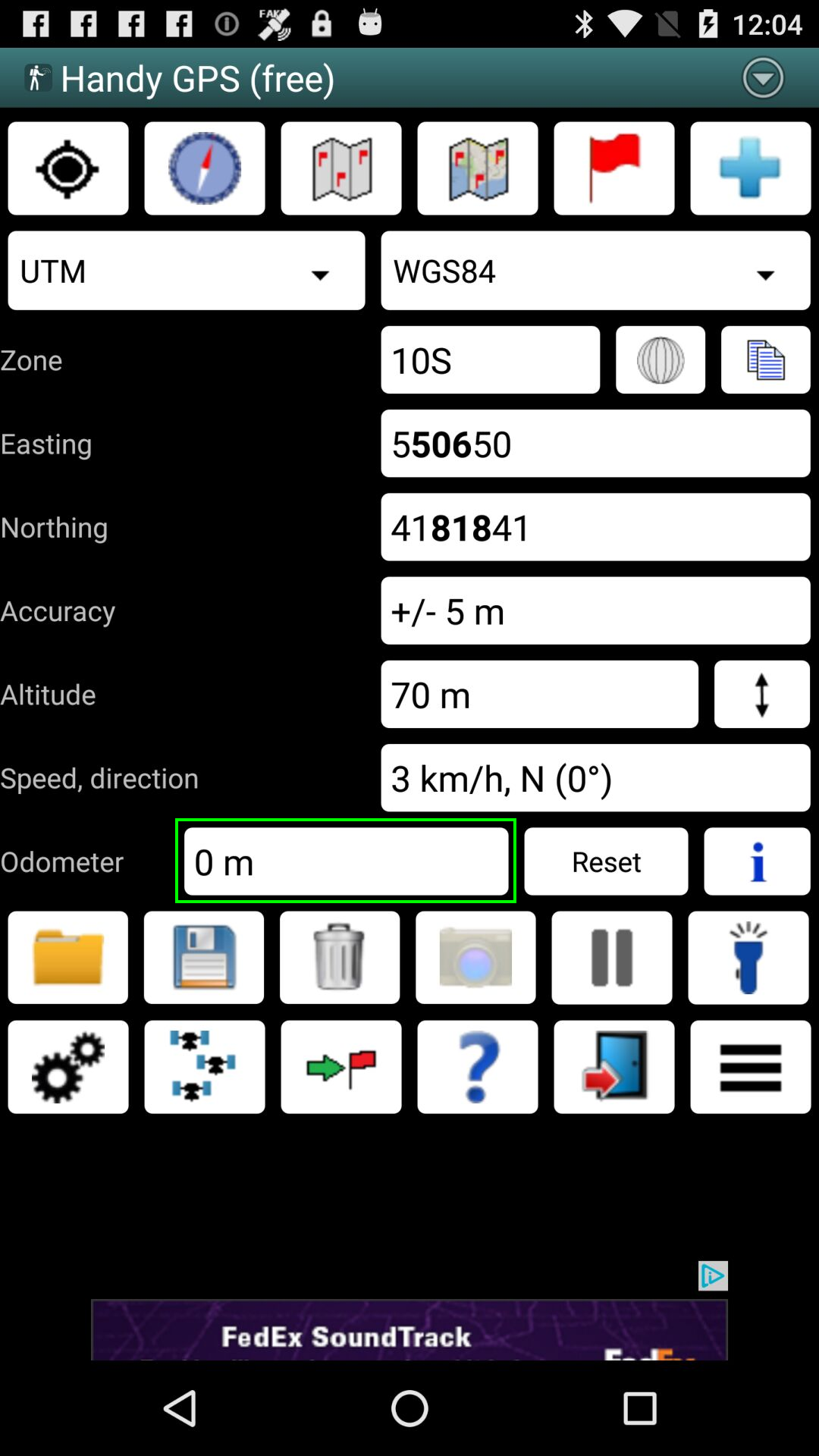}
    }
    & 
    \begin{minipage}{0.65\textwidth}
        \begin{tabular}{ll}
            Question & 
                What is the odometer reading?
            \\ \\
            UI Elem. & 
                \makecell[tl]{ 
                    •\; {[0 m]}
                    •\; {[0 m]}
                } 
            \\ \\
            Long Ans. &
                \makecell[tl]{ 
                    •\; The odometer reading is 0m. \\
                    •\; The odometer reading is 0 m.
                }
            \\ \\
            \makecell[tl]{Generated\\ Short Ans.} &
                \makecell[tl]{
                    •\; 0 m \\
                    •\; 0 meters
                } 
            \\ \\
        \end{tabular}
    \end{minipage} \\
    \midrule
    \raisebox{-33mm}{\includegraphics[height=77mm]{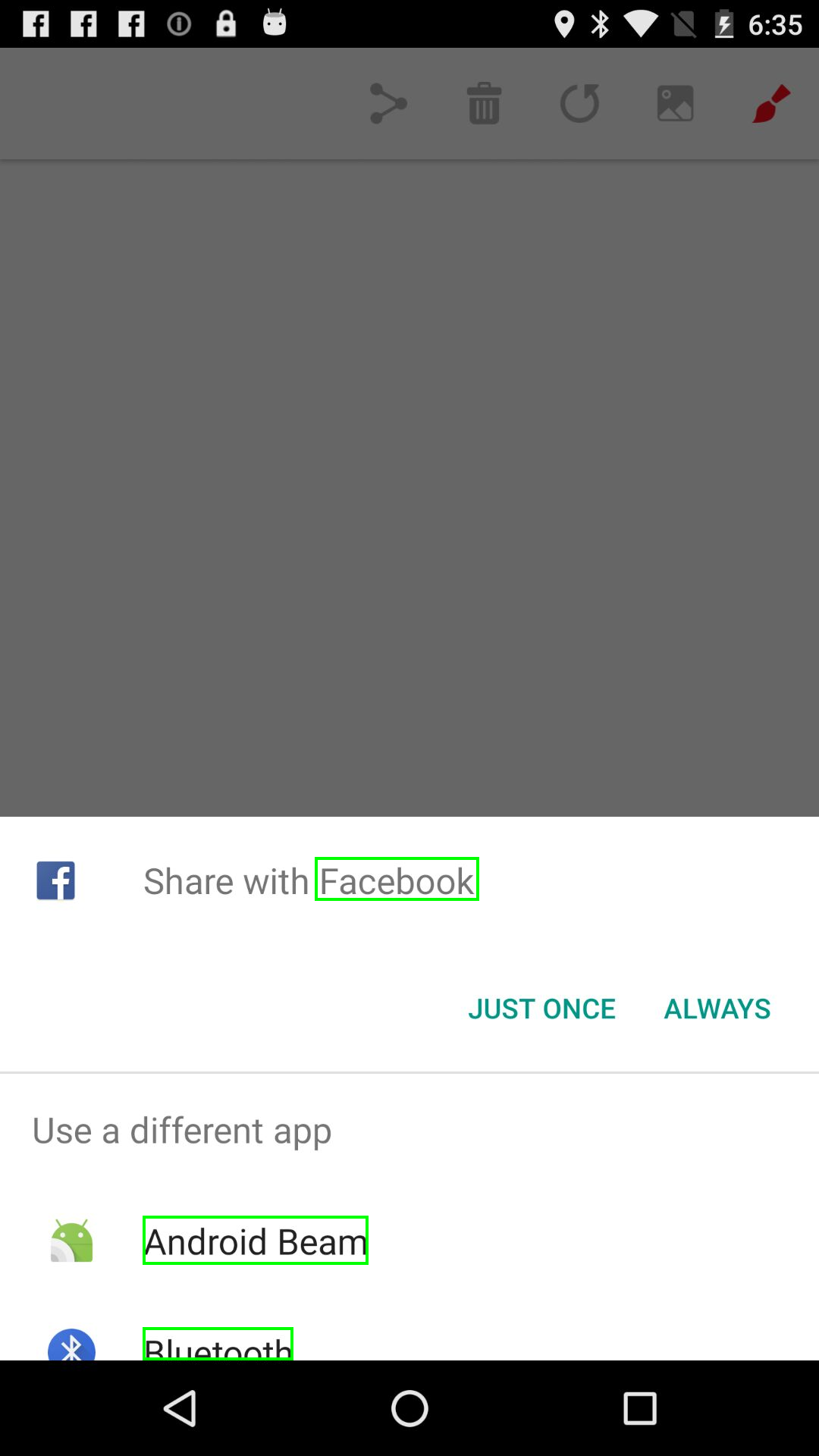}}
    & 
    \begin{minipage}{0.65\textwidth}
        \begin{tabular}{ll}
            Question & What other applications can be used?
            \\ \\
            UI Elem. & 
                \makecell[tl]{
                    •\; {[Android Beam]}, {[Bluetooth]} \\
                    •\; {[Facebook]}, {[Android Beam]}
                } 
            \\ \\
            Long Ans. &
                \makecell[tl]{ 
                    •\; The applications that can be used are "Facebook", \\ \quad "Android Beam", and "Bluetooth". \\
                    •\; The other applications that can be used are \\ \quad "Android Beam" and "Bluetooth".
                }
            \\ \\
            \makecell[tl]{Generated\\ Short Ans.} &
                \makecell[tl]{
                    •\; Android Beam, Bluetooth \\
                    •\; Android Beam and Bluetooth \\
                    •\; "Android Beam", "Bluetooth" \\
                    •\; "Android Beam" and "Bluetooth" \\
                    •\; Facebook, Android Beam, Bluetooth \\
                    •\; Facebook and Android Beam and Bluetooth \\
                    •\; Facebook or Android Beam or Bluetooth
                } 
            \\
        \end{tabular}
    \end{minipage} \\       
    \bottomrule
\caption{Examples from ScreenQA dataset}
\label{tab:examples1}
\end{longtable}
\clearpage
\twocolumn

\begin{figure*}[ht]
    \centering
    \includegraphics[width=0.85\textwidth]{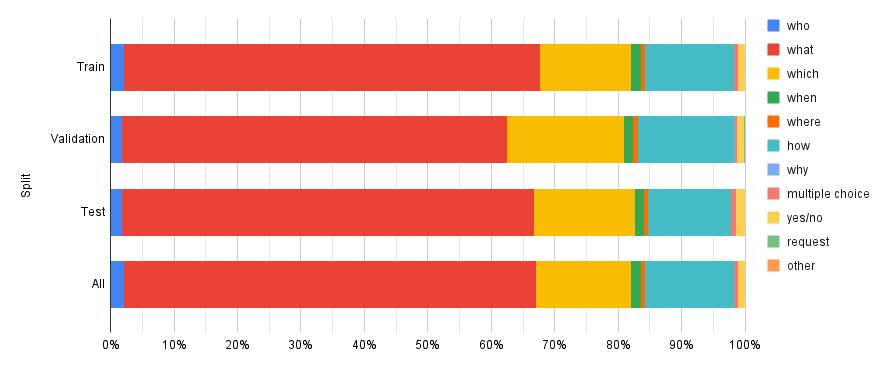}
    \caption{
        Histogram for the types of questions.
    }
    \label{fig:q_types}
\end{figure*}

\section{Dataset Annotation Analysis}
\label{appendix:dataset_analysis}

\begin{table*}[t]
    \centering
    \scriptsize
    \begin{tabular}{lrll}
        \toprule
        Category & \% & \multicolumn{2}{l}{Examples} \\
        \cmidrule(lr){1-2} 
        Subtotal from Table~\ref{tab:top_q_categories} & 83.8 & & \\
        \cmidrule(lr){1-2} \cmidrule(lr){3-4}
        Signup/login & 1.6 & Which application can be used to sign up / login? & What are the alternative choices for signing up? \\
        Version information & 1.6 & What is the version number? & What is the new feature in version v3.1.3? \\
        Weather & 1.5 & What is the range of temperature shown on Sunday? & What is the weather forecast for Sunday? \\
        Score \& value & 1.4 & What is height/weight of the person? & What is the score? \\
        Yes/No & 1.1 & Is there any travel plans? & Is there any favorite? \\
        Phone number & 1.0 & What is the phone number? & What is the prefix for the international mobile number? \\
        \# of Stars & 0.8 & What is the star rating? & How many stars are given to the product? \\
        Share/sharing & 0.8 & Which application can be used to share? & Where can I share this application? \\
        Age & 0.8 & How old is ...? & What is the age? \\
        Percentage & 0.7 & What is the percentage of ... ? & What is the brightness percentage for foreground? \\
        Settings & 0.6 & What is the setting of ... ? & Which settings are switched on? \\
        Quantity amount & 0.6 & How much fat is there? & What is the amount? \\
        Permission & 0.5 & Which application is asking for permissions? & What permissions are required for MyCarTracks? \\
        \# of Likes & 0.5 & How many likes for ... ? & How many likes does ... get? \\
        Country & 0.5 & What is the name of the country? & Which country has the +54 code? \\
        Distance & 0.5 & What is the visibility distance? & How far is it from ... ? \\
        \# of Reviews & 0.4 & What is the number of comments on ... ? & How many comments? \\
        Website & 0.3 & What is the url? & What's the website address? \\
        Gender & 0.3 & What is the gender? & Which gender is displayed on the screen? \\
        How to & 0.3 & How to start on boot? & How to pronounce his name? \\
        Currency & 0.3 & What is the currency? & What is the currency for the price? \\
        Unit of measurement & 0.2 & What is the unit of temperature? & What is the unit of weight and length? \\
        Language & 0.1 & Which language is used in the setting? & Which language is being translated into which language? \\
        Color & 0.0 & What is the UI color? & What is the amount of green color? \\
        \cmidrule(lr){1-2}
        Total & 100.0 & & \\
    \bottomrule
    \end{tabular}
    \caption{Remaining question categories and examples (\textit{cont.} from Table~\ref{tab:top_q_categories})}
    \label{tab:full_q_categories}
\end{table*}

Additional dataset annotation analysis is provided in this section.
Table~\ref{tab:full_q_categories} shows the remaining question categories and examples continuing from Table~\ref{tab:top_q_categories}.
Figure~\ref{fig:q_types} shows distribution of question types (e.g., Wh- questions, Yes/No questions, etc.), regardless of the subject.

\section{Evaluation Configurations}
\label{appendix:model_evaluation_details}

In Section~\ref{sec:baselines} we report the performance of various models on ScreenQA tasks in a zero-shot setting and after fine-tuning. We provide additional details as to how we evaluated the baselines for the corresponding model sizes.

\subsection{Zero-shot}
\label{appendix:zero_shot_evaluation_details}

Question answering is one of the most common tasks for LLMs and VLMs.
Since the output format of the \textit{SQA-S} task is similar to other existing benchmarks, we attempt evaluating publicly available models in zero-shot setting.
Specifically, we focus our evaluations on Fuyu-8b\footnote{\href{https://www.adept.ai/blog/fuyu-8b/}{Fuyu-8b}}~\cite{fuyu-8b}, Gemini 1.5 Flash\footnote{\href{https://deepmind.google/technologies/gemini/flash/}{Gemini 1.5 Flash}}, Gemini 1.5 Pro\footnote{\href{https://deepmind.google/technologies/gemini/pro/}{Gemini 1.5 Pro}} \cite{team2023gemini}, and GPT-4o\footnote{\href{https://openai.com/index/hello-gpt-4o/}{GPT-4o}}~\cite{openai2024gpt4}.
We refer to Table~\ref{tab:zero_shot_sqas_results} for the results.

Further, we describe for each model the prompt used in the evaluation is an instruction followed by a question.
Fuyu-8B model came with an author recommendation for a prompt, which we made use of \footnote{See \href{https://huggingface.co/adept/fuyu-8b/discussions/28}{Fuyu-8B discussion}}:
\begin{lstlisting}
Answer the following DocVQA question based on the image. \n 
\end{lstlisting}

For GPT-4o we optimized the prompt using 10 examples from the validation split. The one we picked was:
\begin{lstlisting}
Answer the question based on the screenshot only. Do not use any other sources of information. The answer should be succinct and as short as possible. If the answer is a text from the image, provide it exactly without rephrasing or augmenting. If there is no answer on the image, output "<no answer>".\n
\end{lstlisting}

The same prompt was re-used for the Gemini model evaluation. Although this may represent a disadvantage for Gemini compared to GPT-4o, our results indicate that Gemini 1.5 Flash is on par with and 1.5 Pro outperforms GPT-4o using it.

\subsection{Fine-tuning}
\label{appendix:model_finetuning_details}

The training data for SQA-L, SQA-UIC and SQA-UIC-BB tasks has one annotation per example (with a few exceptions), so it's usage for fine-tuning is straightforward for all models.
Short answers, on the other hand, were generated automatically (see Section~\ref{sec:short_answers_generation}), resulting in multiple answers per question for the SQA-S task.
The fine-tuning setup for PaliGemma~3B~\cite{beyer2024paligemma} and ScreenAI~5B~\cite{baechler2024screenai} models randomly select one answer each time. Through sufficient epoch, it is however likely that all answer variants are utilized.

For Gemini 1.5 Flash fine-tuning on SQA-S task, however, one answer per question was randomly selected before training, limiting the diversity of answers observed by the model.
We therefore ran fine-tuning twice for 2 different random selections, and reported the best obtained results among the two experiments.

\section{Observed prediction errors on SQA-S task}
\label{appendix:observed_prediction_errors}

While Section~\ref{sec:baselines} contains evaluations of different models in different settings to provide baselines, in this paper we deliberately refrain from making conclusions about upsides and downsides of each individual model.
Instead, in Table~\ref{tab:observed_error_examples} we provide examples from SQA-S task of some errors models made during evaluation, as a demonstration of complexity of this dataset.
While this may not be a complete list, we categorized those errors into:
\begin{itemize}
    \item \textit{Misinterpreted question}: the predicted answer answers a different question.
    \item \textit{Misinterpreted screenshot info}: the predicted answer is based on the screenshot information that has a different meaning.
    \item \textit{Hallucination}: the predicted answer is not based on the screen at all.
    \item \textit{Modified or misread content}: the predicted answer originates from correct answer but contains typos or is incorrectly transformed into something different.
    \item \textit{Lack of understanding/reasoning to identify/compose answer}: the predicted answer suggests poor model's screen understanding or reasoning capabilities.
\end{itemize}

\clearpage
\onecolumn
\begin{longtable}{ll}
    \toprule
    \raisebox{-35mm}{\includegraphics[height=75mm]{figures/screenshots_error_analysis/rico_com.awestruckstudios.pkmn_0_546.png}}
    & 
    \begin{minipage}{0.65\textwidth}
        \begin{tabular}{p{0.25\linewidth} p{0.7\linewidth}}
            Question & What is the total number?
            \\ \\
            Correct answer & 600
            \\ \\
            Incorrect answer & 721
            \\ \\
            Error category & Misinterpreted question
            \\ \\
            Explanation & The serial number present on the screen was incorrectly identified as the total.
            \\ \\
        \end{tabular}
    \end{minipage} \\
    \midrule
    \raisebox{-35mm}{\includegraphics[height=75mm]{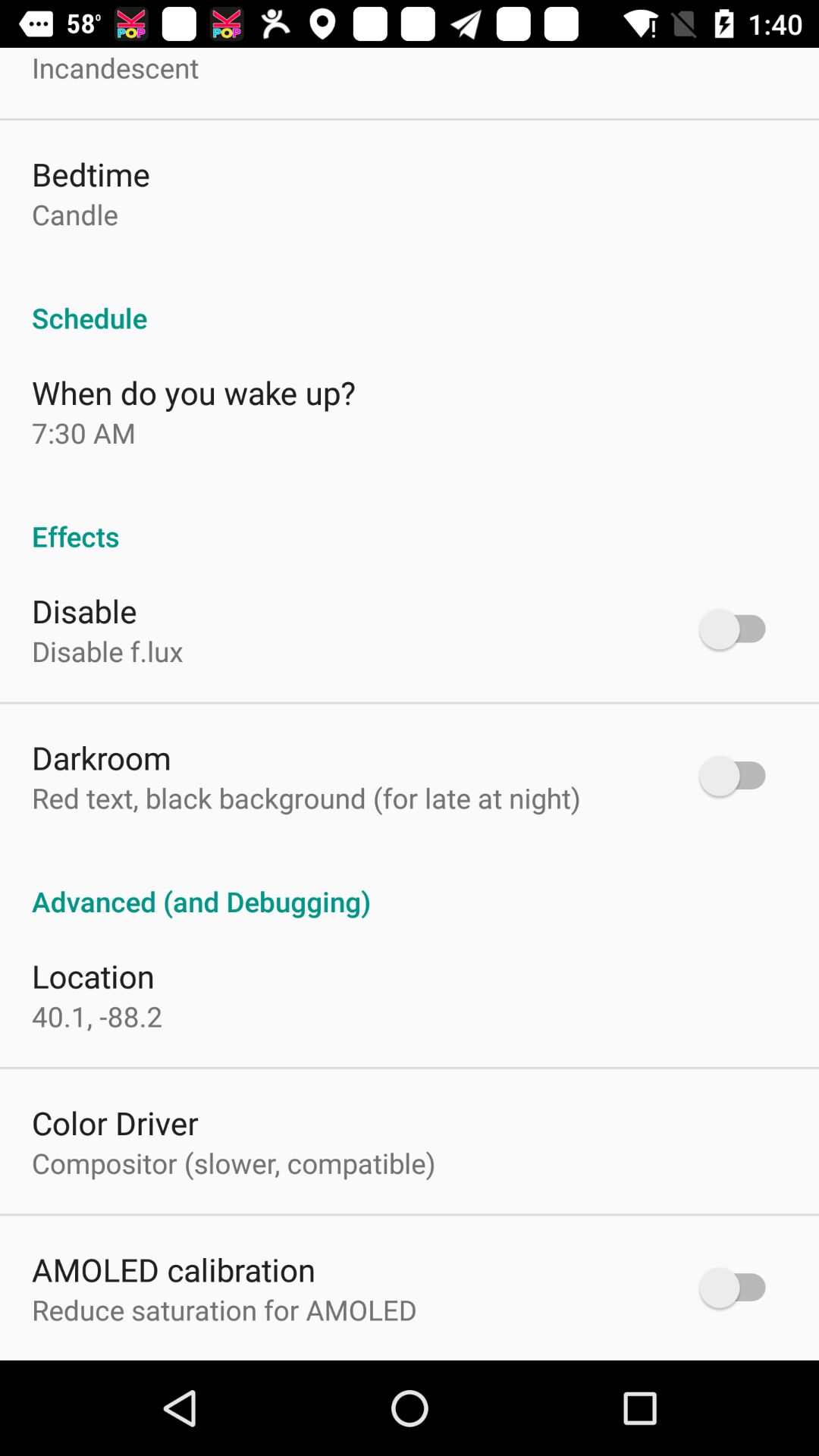}}
    & 
    \begin{minipage}{0.65\textwidth}
        \begin{tabular}{p{0.25\linewidth} p{0.7\linewidth}}
            Question & For which setting is the "Compositor" option selected?
            \\ \\
            Correct answer & "Color Driver" setting
            \\ \\
            Incorrect answer & Advanced (and Debugging)
            \\ \\
            Error category & Misinterpreted question
            \\ \\
            Explanation & "Advanced (and Debugging)" is a group of settings, and "Color Driver" is a single setting in that group, which is the correct answer.
            \\ \\
        \end{tabular}
    \end{minipage} \\
    \midrule
    \raisebox{-35mm}{\includegraphics[height=75mm]{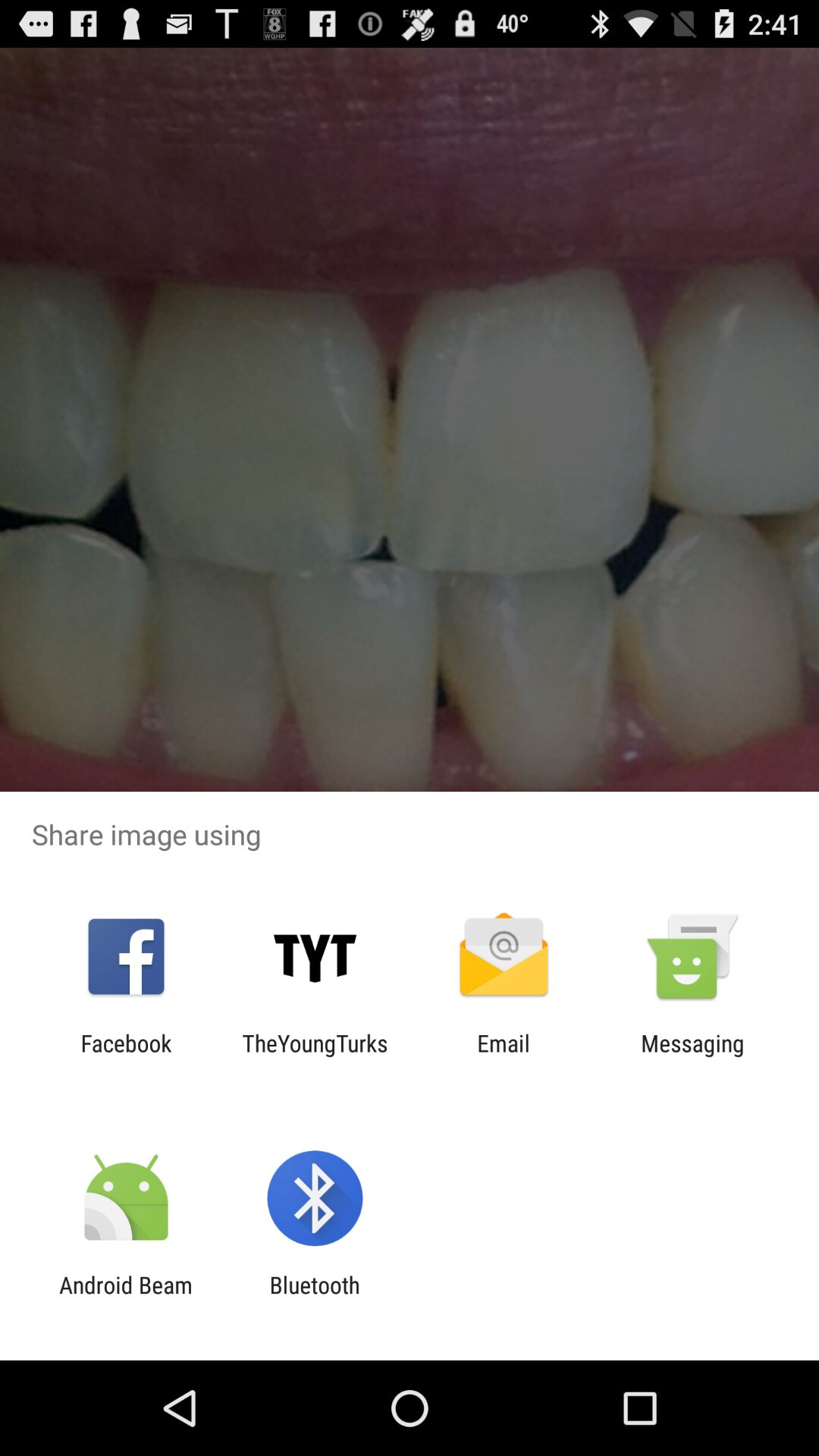}}
    & 
    \begin{minipage}{0.65\textwidth}
        \begin{tabular}{p{0.25\linewidth} p{0.7\linewidth}}
            Question & Through which applications can we open the browser?
            \\ \\
            Correct answer & \(<\)no answer\(>\)
            \\ \\
            Incorrect answer & Facebook, TheYoungTurks, Email, Messaging, Android Beam and Bluetooth
            \\ \\
            Error category & Misinterpreted question
            \\ \\
            Explanation & Predicted the ways to share content, instead of opening the browser.
            \\ \\
        \end{tabular}
    \end{minipage} \\
    \midrule
    \raisebox{-35mm}{\includegraphics[height=75mm]{figures/screenshots_error_analysis/rico_com.blogspot.fuelmeter_0_297.png}}
    & 
    \begin{minipage}{0.65\textwidth}
        \begin{tabular}{p{0.25\linewidth} p{0.7\linewidth}}
            Question & What's the title of the expense amount for February 15th?
            \\ \\
            Correct answer & Other
            \\ \\
            Incorrect answer & 50000.00 USD
            \\ \\
            Error category & Misinterpreted screenshot info
            \\ \\
            Explanation & The expense amount of 50000.0 USD  was mistakenly identified as the title.
            \\ \\
        \end{tabular}
    \end{minipage} \\
    \midrule
    \raisebox{-35mm}{\includegraphics[height=75mm]{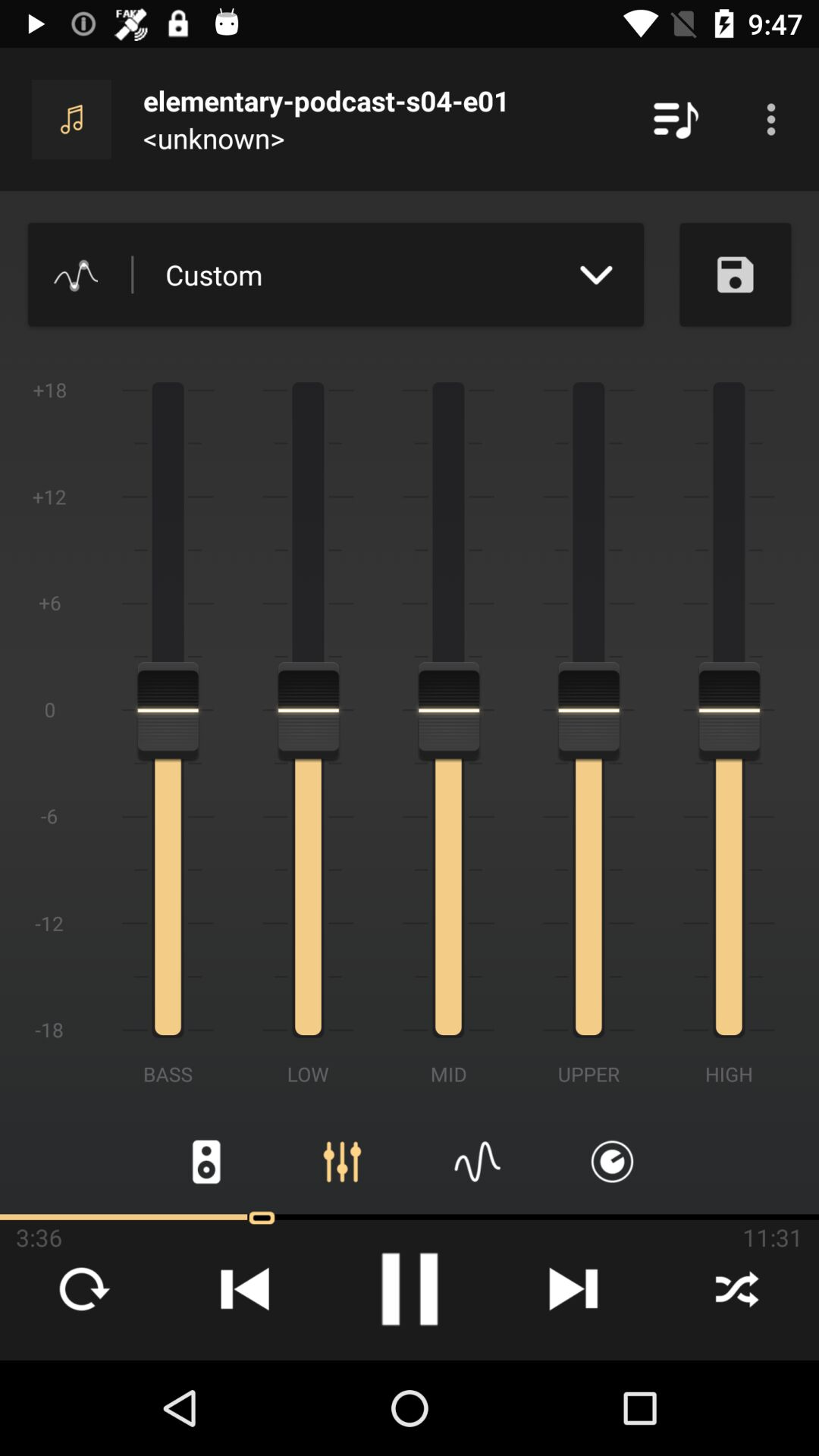}}
    & 
    \begin{minipage}{0.65\textwidth}
        \begin{tabular}{p{0.25\linewidth} p{0.7\linewidth}}
            Question & How long is the audio?
            \\ \\
            Correct answer & 11 minutes and 31 seconds
            \\ \\
            Incorrect answer & 3:36
            \\ \\
            Error category & Misinterpreted screenshot info
            \\ \\
            Explanation & The current audio playing position at 3:36 was identified as the total duration instead of 11:31.
            \\ \\
        \end{tabular}
    \end{minipage} \\
    \midrule
    \raisebox{-35mm}{\includegraphics[height=75mm]{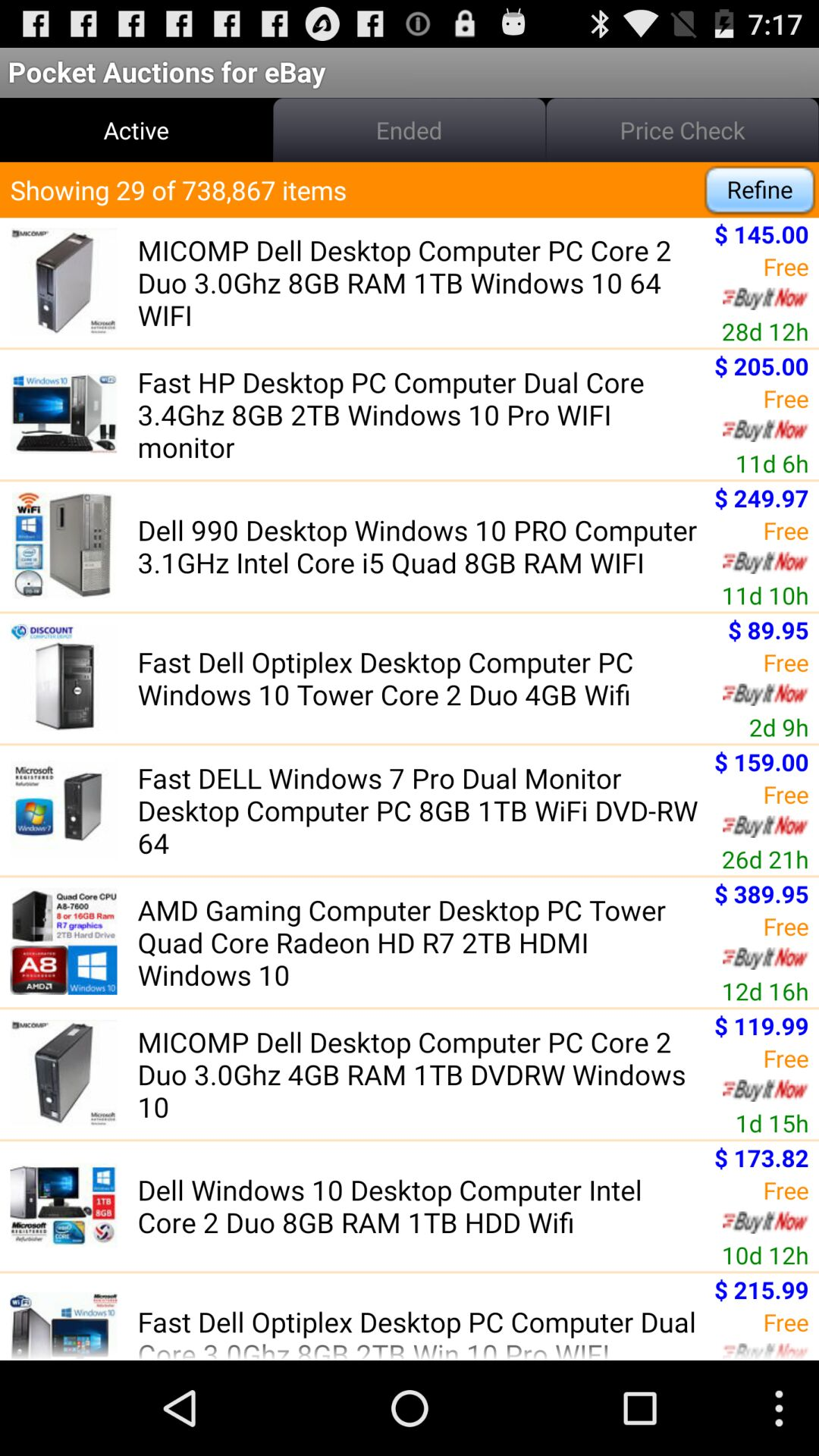}}
    & 
    \begin{minipage}{0.65\textwidth}
        \begin{tabular}{p{0.25\linewidth} p{0.7\linewidth}}
            Question & What is the number of shown items?
            \\ \\
            Correct answer & 29
            \\ \\
            Incorrect answer & 738,867
            \\ \\
            Error category & Misinterpreted screenshot info
            \\ \\
            Explanation & The total number of items is 738,867, while 29 is the shown number.
            \\ \\
        \end{tabular}
    \end{minipage} \\
    \midrule
    \raisebox{-35mm}{\includegraphics[height=75mm]{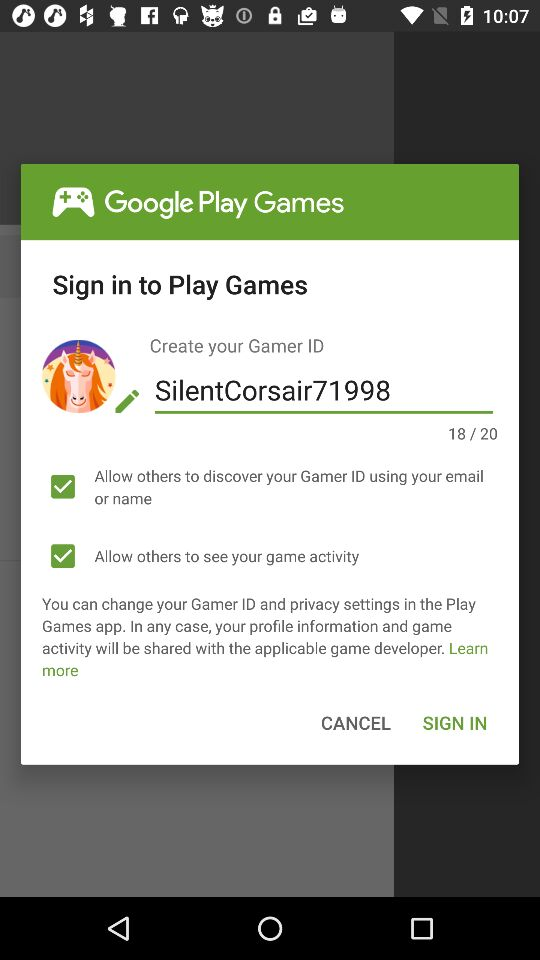}}
    & 
    \begin{minipage}{0.65\textwidth}
        \begin{tabular}{p{0.25\linewidth} p{0.7\linewidth}}
            Question & When was the article published?
            \\ \\
            Correct answer & \(<\)no answer\(>\)
            \\ \\
            Incorrect answer & 18 / 20
            \\ \\
            Error category & Misinterpreted screenshot info
            \\ \\
            Explanation & 18 / 20 represents the number of characters in a text field and not the date.
            \\ \\
        \end{tabular}
    \end{minipage} \\
    \midrule
    \raisebox{-35mm}{\includegraphics[height=75mm]{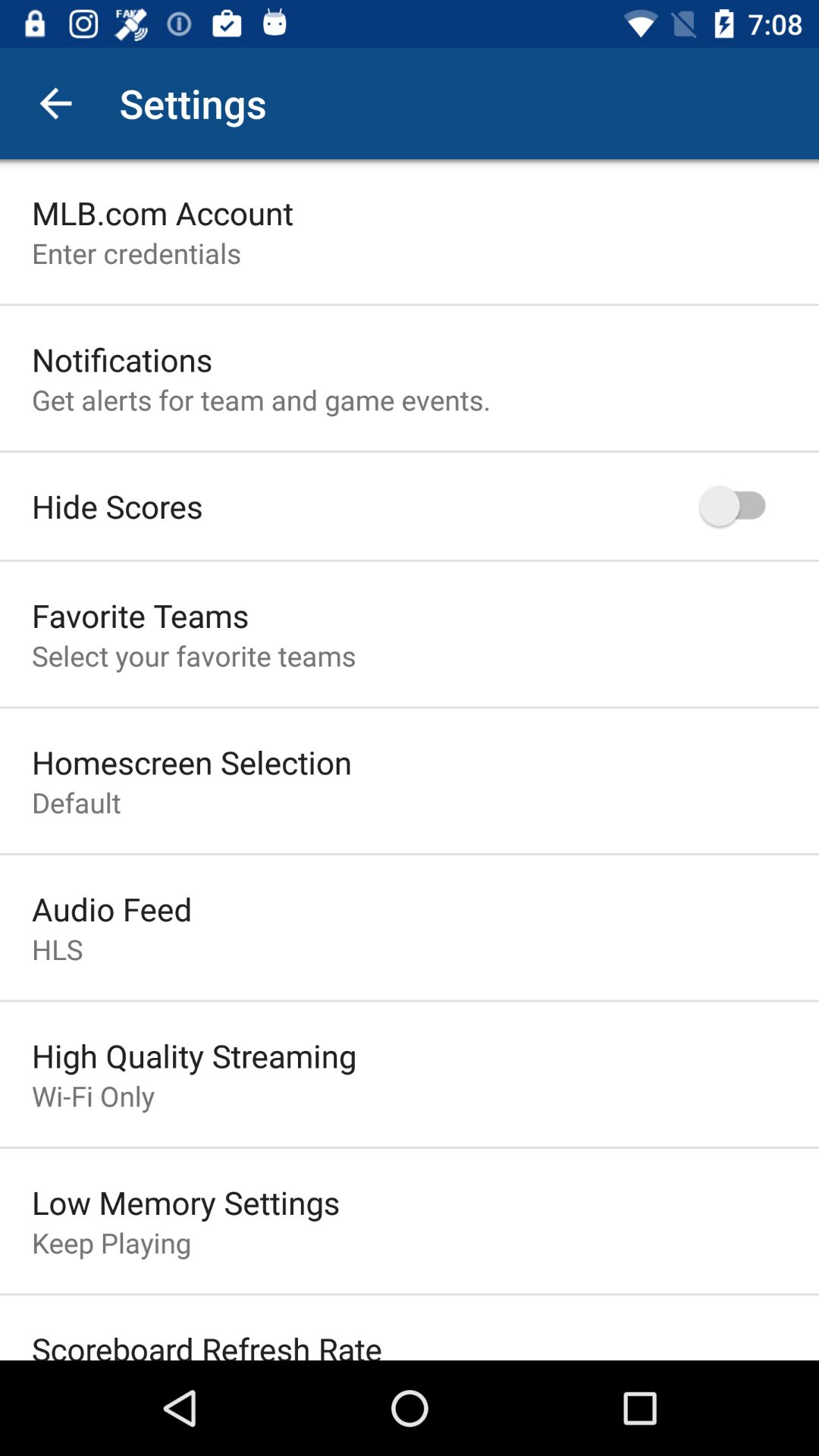}}
    & 
    \begin{minipage}{0.65\textwidth}
        \begin{tabular}{p{0.25\linewidth} p{0.7\linewidth}}
            Question & Which option is selected for high quality streaming?
            \\ \\
            Correct answer & Wi-Fi Only
            \\ \\
            Incorrect answer & "Only for VIPs" option
            \\ \\
            Error category & Hallucination
            \\ \\
            Explanation & "High Quality Streaming" setting has value "Wi-Fi Only".
            \\ \\
        \end{tabular}
    \end{minipage} \\
    \midrule
    \raisebox{-35mm}{\includegraphics[height=75mm]{figures/screenshots_error_analysis/rico_com.divum.CricketNext_0_584.png}}
    & 
    \begin{minipage}{0.65\textwidth}
        \begin{tabular}{p{0.25\linewidth} p{0.7\linewidth}}
            Question & Who has taken six wickets?
            \\ \\
            Correct answer & Yuzvendra Chahal
            \\ \\
            Incorrect answer & Lelebhim Yuzvendra Chahal
            \\ \\
            Error category & Hallucination
            \\ \\
            Explanation & Possibly the word "Leg-spinner" was misread or hallucinated into "Lelebhim".
            \\ \\
        \end{tabular}
    \end{minipage} \\
    \midrule
    \raisebox{-35mm}{\includegraphics[height=75mm]{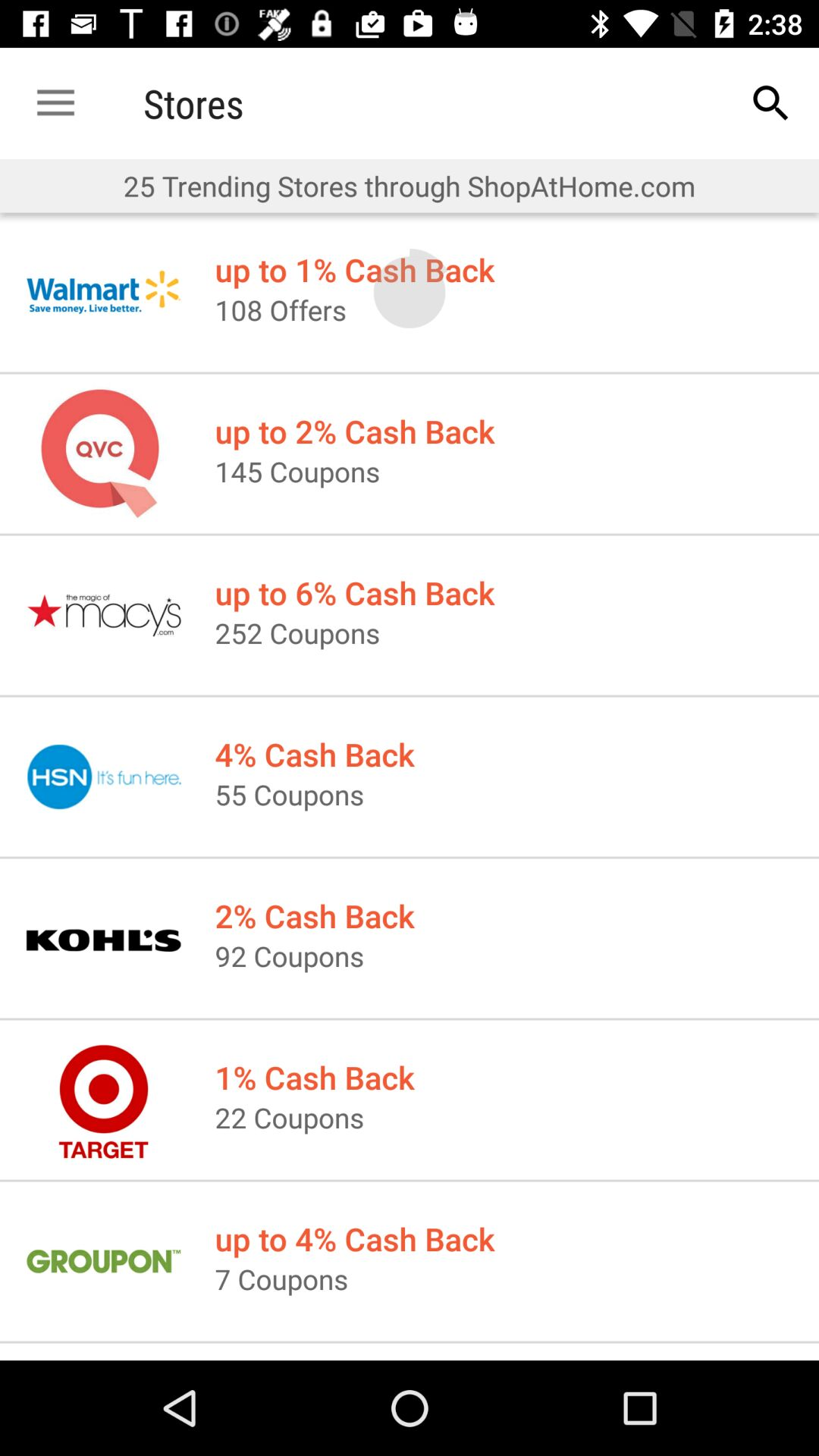}}
    & 
    \begin{minipage}{0.65\textwidth}
        \begin{tabular}{p{0.25\linewidth} p{0.7\linewidth}}
            Question & Which company is giving 252 coupons?
            \\ \\
            Correct answer & macy's
            \\ \\
            Incorrect answer & "mady's"
            \\ \\
            Error category & Modified or misread content
            \\ \\
            Explanation & "Macy's" was misspelled.
            \\ \\
        \end{tabular}
    \end{minipage} \\
    \midrule
    \raisebox{-35mm}{\includegraphics[height=75mm]{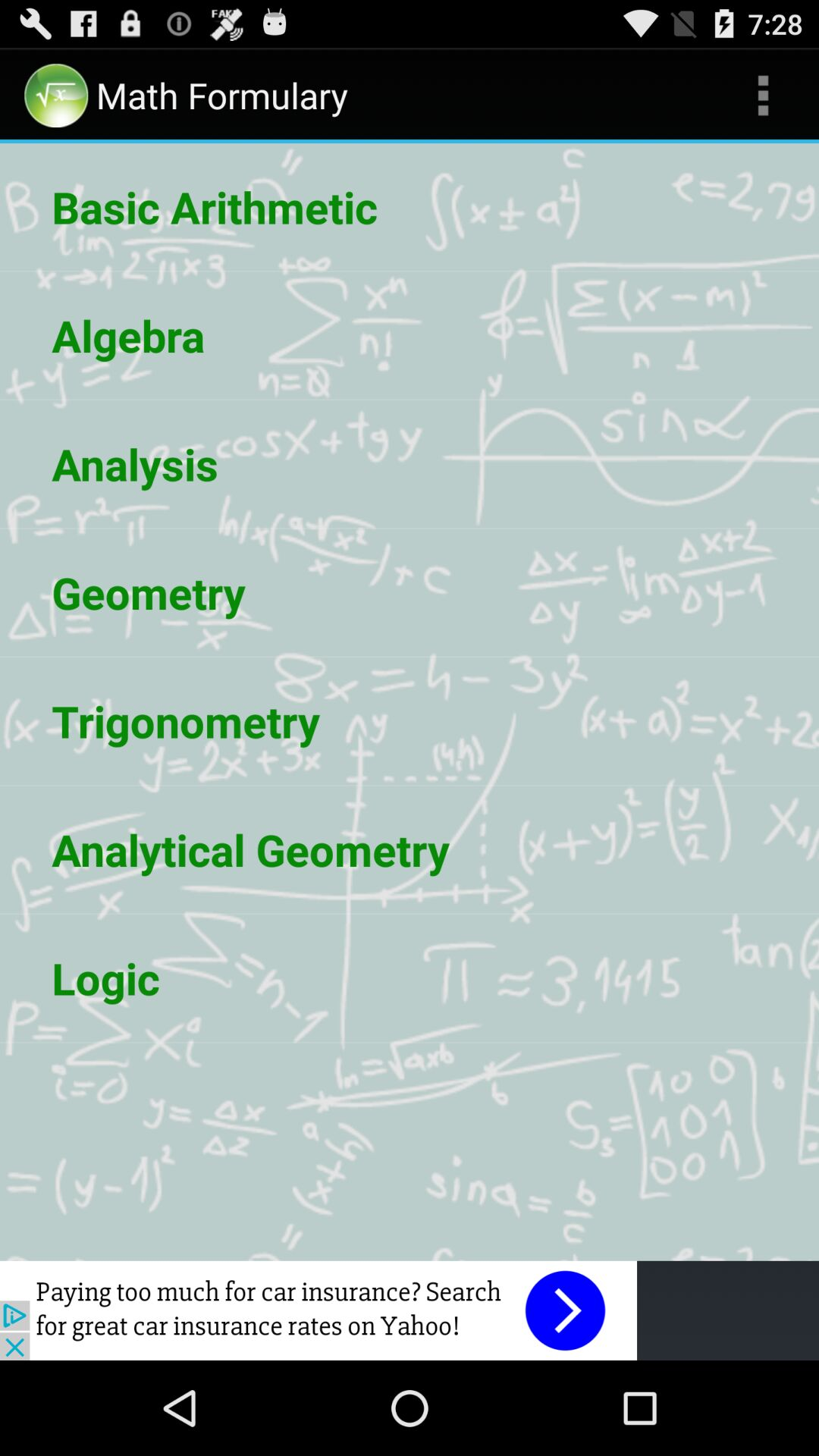}}
    & 
    \begin{minipage}{0.65\textwidth}
        \begin{tabular}{p{0.25\linewidth} p{0.7\linewidth}}
            Question & What is the application name?
            \\ \\
            Correct answer & Math Formulary
            \\ \\
            Incorrect answer & Math Formulas Free
            \\ \\
            Error category & Modified or misread content
            \\ \\
            Explanation & Hallucination of "Forumlas Free" instead of "Formulary"
            \\ \\
        \end{tabular}
    \end{minipage} \\
    \midrule
    \raisebox{-35mm}{\includegraphics[height=75mm]{figures/screenshots_error_analysis/rico_com.freedomlabs.free.music.archive_0_2128.png}}
    & 
    \begin{minipage}{0.65\textwidth}
        \begin{tabular}{p{0.25\linewidth} p{0.7\linewidth}}
            Question & What are the different types of genres?
            \\ \\
            Correct answer & "AVANT-GARDE", "INTERNATIONAL", "BLUES", "JAZZ", "CLASSICAL" and "NOVELTY"
            \\ \\
            Incorrect answer & "AVANT-garde", "INTERNATIONAL", "BLUES", "JAZZ", "CLASSICAL" and "SAVELY"
            \\ \\
            Error category & Modified or misread content
            \\ \\
            Explanation & Hallucination of "SAVELY" instead of "NOVELTY"
            \\ \\
        \end{tabular}
    \end{minipage} \\
    \midrule
    \raisebox{-35mm}{\includegraphics[height=75mm]{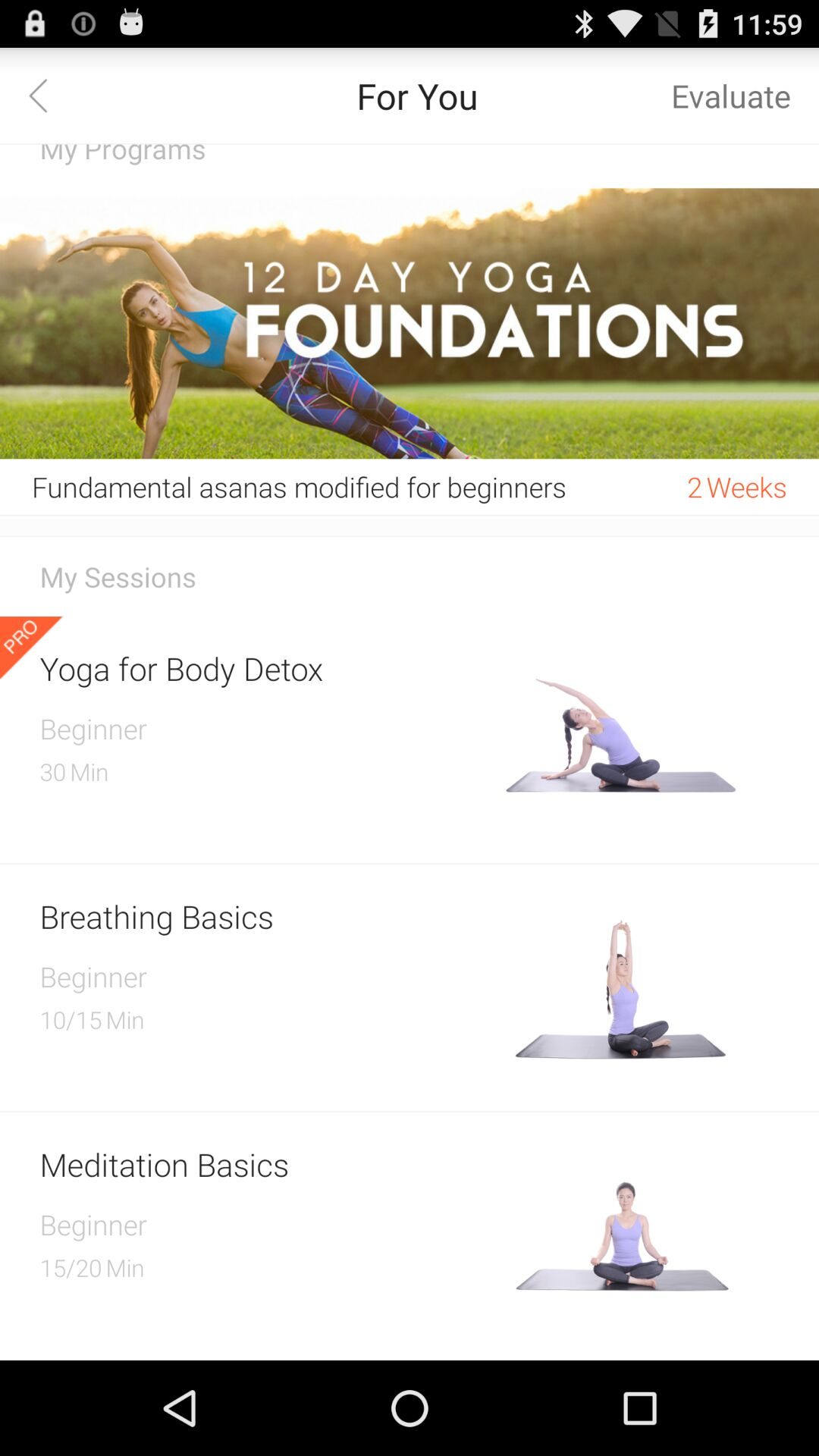}}
    & 
    \begin{minipage}{0.65\textwidth}
        \begin{tabular}{p{0.25\linewidth} p{0.7\linewidth}}
            Question & How many minutes does a beginner have to do the breathing basics session?
            \\ \\
            Correct answer & 10 to 15 minutes
            \\ \\
            Incorrect answer & 10 minutes and 15 minutes
            \\ \\
            Error category & Lack of understanding/reasoning to identify/compose answer
            \\ \\
            Explanation & The UI indicates that the breathing session is "10/15 min", which from the context should be interpreted as 10 to 15 mins, rather than 10 and 15 mins, separately.
            \\ \\
        \end{tabular}
    \end{minipage} \\
    \midrule
    \raisebox{-35mm}{\includegraphics[height=75mm]{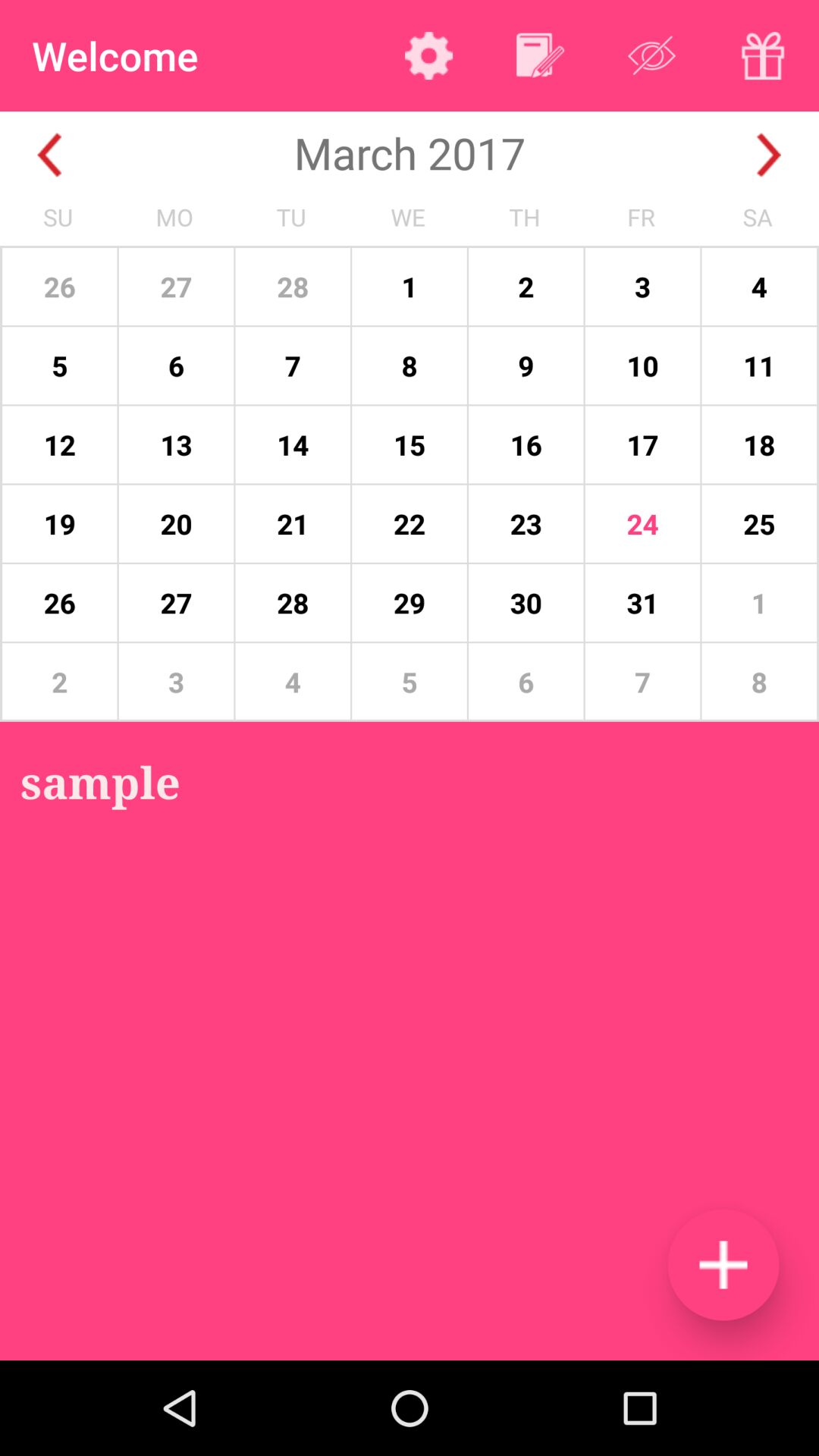}}
    & 
    \begin{minipage}{0.65\textwidth}
        \begin{tabular}{p{0.25\linewidth} p{0.7\linewidth}}
            Question & What is the day on the 17th of March?
            \\ \\
            Correct answer & FR
            \\ \\
            Incorrect answer & TH
            \\ \\
            Error category & Lack of understanding/reasoning to identify/compose answer
            \\ \\
            Explanation & The day of the week was not correctly identified from the calendar.
            \\ \\
        \end{tabular}
    \end{minipage} \\
    \bottomrule
    \caption{Examples of errors made by the models from ScreenQA dataset on SQA-S task.}
    \label{tab:observed_error_examples}
\end{longtable}
\clearpage
\twocolumn

\end{document}